\documentclass[lettersize,journal]{IEEEtran}
\usepackage{amsmath,amsfonts}
\usepackage{algorithmic}
\usepackage{algorithm}
\usepackage{array}
\usepackage[caption=false,font=normalsize,labelfont=sf,textfont=sf]{subfig}
\usepackage{stfloats}
\usepackage{url}
\usepackage{verbatim}
\usepackage{graphicx}
\usepackage{cite}
\usepackage{gensymb}
\usepackage{multirow}
\usepackage{ulem}
\usepackage{rotating}
\usepackage{makecell}
\usepackage{tikz}
\usetikzlibrary{spy}
\usepackage{adjustbox}
\usepackage{hyperref}

\usepackage{stix}
\usepackage{chemarrow}

\usepackage{color, xcolor}

\begin{document}

\title{Relighting from a Single Image: Datasets and Deep Intrinsic-based Architecture}

\author{Yixiong Yang, Hassan Ahmed Sial, Ramon Baldrich, Maria Vanrell
        % <-this % stops a space
\vspace{-10pt}
\thanks{This paper was produced by Yixiong Yang, Ramon Baldrich, Maria Vanrell (Corresponding Author) from Computer Vision Center, Universitat Autònoma de Barcelona, and Hassan Ahmed Sial from ISGlobal. They are in Barcelona, Spain.}% <-this % stops a space
\thanks{Manuscript submitted to IEEE Transactions on Multimedia; }
}

% The paper headers
% \markboth{Journal of \LaTeX\ Class Files,~Vol.~14, No.~8, July~2023}%
\markboth{Journal of \LaTeX\ Class Files}
{Shell \MakeLowercase{\textit{et al.}}: A Sample Article Using IEEEtran.cls for IEEE Journals}

\IEEEpubid{0000--0000/00\$00.00~\copyright~2023 IEEE}

\maketitle

\begin{abstract}
Single image scene relighting aims to generate a realistic new version of an input image so that it appears to be illuminated by a new target light condition. Although existing works have explored this problem from various perspectives, generating relit images under arbitrary light conditions remains highly challenging, and related datasets are scarce. Our work addresses this problem from both the dataset and methodological perspectives. We propose two new datasets: a synthetic dataset with the ground truth of intrinsic components and a real dataset collected under laboratory conditions. These datasets alleviate the scarcity of existing datasets. To incorporate physical consistency in the relighting pipeline, we establish a two-stage network based on intrinsic decomposition, giving outputs at intermediate steps, thereby introducing physical constraints. When the training set lacks ground truth for intrinsic decomposition, we introduce an unsupervised module to ensure that the intrinsic outputs are satisfactory. Our method outperforms the state-of-the-art methods in performance, as tested on both existing datasets and our newly developed datasets. Furthermore, pretraining our method or other prior methods using our synthetic dataset can enhance their performance on other datasets.  Since our method can accommodate any light conditions, it is capable of producing animated results. The dataset, method, and videos are publicly available at \url{https://github.com/CVC-CIC/DeepIntrinsicRelighting}. 

\end{abstract}
\vspace{-1pt}

\vspace{-3pt}
\section{Introduction}
\vspace{-1pt}

The task of image relighting involves generating a revised version of an original image such that it appears to be illuminated realistically under a given new light condition. In recent years, relighting has emerged as an essential component in a range of fields, including augmented reality (AR), professional photography for aesthetic enhancement, and photo montage \cite{helou2020vidit, nestmeyer2020learning, caselles2023sira, zhu2022ian, zhang2021pr, han2019asymmetric}. 

While the topic of relighting is not new in the field of computer vision \cite{moreno2005optimal}, it remains an ongoing challenge. The development of deep learning has provided many new ways to solve the relighting problem. Many researchers have developed different methods, designed to handle relighting in various scenarios and with different kinds of inputs. For example, \cite{nestmeyer2020learning,sun2019single,zhou2019deep,pandey2021total} focused on portrait relighting using a single facial image as input, whereas others \cite{duchene2015multi,xu2018deep,philip2019multi, moreno2005optimal, srinivasan2021nerv} have performed relighting using multiple images. Despite these advancements, the challenge of relighting a single image to any target light condition remains, especially when the content encompasses an entire scene rather than a single object. This is precisely the issue we aim to address in this paper. 

Two datasets related to our objective have been proposed. Murmann \textit{et al.} \cite{murmann2019dataset} created an indoor real-world multi-illumination dataset where each scene is captured under varying light conditions. Helou \textit{et al.} \cite{helou2020vidit} proposed the synthetic VIDIT dataset, with each scene synthesized under different positions and temperatures of light. Based on this dataset, the authors also organized several relighting challenges \cite{helou2020aim, el2021ntire}. These two datasets form the basis for various studies \cite{zhu2022ian, kubiak2021silt}.
Despite the significant and commendable progress made by these two datasets, there is scope for improvement owing to the ill-posed nature of relighting. The Multi-Illumination dataset uses the pictures of two probes to represent the light condition, thus limiting the ability of the user to freely input light conditions during testing.  In the VIDIT dataset, some scenes contain extensive dark areas, introducing significant difficulties for relighting, since these black areas are devoid of any information. Relighting these areas seems to be impossible, and indeed, no methods have been able to accomplish this according to the results from the challenge \cite{helou2020aim}. Therefore, new datasets to enrich the diversity of research in the community are needed.

\IEEEpubidadjcol

Three key factors contribute to the complexity of single image relighting. First, it necessitates the removal of effects caused by the original light condition, specifically the elimination of shading effects and cast shadows. 
Second, it requires inferring the intrinsic shape properties of scene objects from the image. Finally, based on the target light condition and the estimated shape properties, the scene must be rerendered, and new shadows must be appropriately generated. In addition, the task becomes more complex as we transition from focusing on specific scenes, such as faces or individual objects, to broader, more generic scenes. This complexity arises from an increase in the number of objects and the need to account for the interactions of shadows, where the shadow cast from one object may be cast onto another.
The challenge of relighting remains unresolved; existing methods continue to demonstrate significant noise effects in areas where shadows should be removed, where shadows are generated, and along object edges. In response, we draw inspiration from intrinsic decomposition to address these issues \cite{barrow1978recovering}.

From the discussion above, we believe that existing research could be further developed in two primary ways: first, by formulating new datasets that facilitate the training and validation of new methods; and second, by proposing a strategy founded on intrinsic decomposition, which could lend beneficial physical constraints to relighting tasks. With these considerations, in this work, we present three contributions:
\vspace{-2pt}
\begin{itemize}
\item First, we create a comprehensive synthetic dataset specifically designed for relighting tasks, which also includes the ground truth (GT) for intrinsic decomposition. 

\item Second, we construct a real-world dataset within a controlled laboratory setting, to prove that our method can work on real scenes. Moreover, we demonstrate that our knowledge trained on synthetic data can be transferred to real scenes. 

\item Third, we introduce a two-stage network architecture that hinges on intrinsic decomposition. This network is designed to output intrinsic components at an intermediate stage, which allows us to incorporate physical constraints. In addition, we introduce a module that enables unsupervised intrinsic decomposition when the GT of intrinsic decomposition is not available. 
\end{itemize}
\vspace{-7pt}
\section{Related works}
Recently relighting has been explored from different points of view \cite{einabadi2021deep}. In this section, we first discuss the work that is most directly related to our objectives which are from a single image. Then, we provide a brief review of other related studies carried out from divergent perspectives to differentiate them from our goals. Finally, since we have employed intrinsic decomposition, we also summarize the relevant work.
\vspace{-7pt}
\subsection{Scene relighting from a single image}

As previously mentioned, Murmann \textit{et al.} \cite{murmann2019dataset} introduced the Multi-Illumination dataset. They also explored relighting through a deep network, where the input light conditions consisted of nine options while the target was fixed. Helou \textit{et al.} \cite{helou2020vidit} proposed the synthetic VIDIT dataset and hosted the AIM 2020 \cite{helou2020aim,gardner2019deep,boss2020two, puthussery2020wdrn, wang2020deep} and NTIRE 2021 \cite{el2021ntire} challenges. Given that the latter requires depth during inference, we are more interested in the former because of its wider range of application scenarios. This challenge \cite{helou2020aim} featured two distinct tracks: one-to-one and any-to-any relighting, where "one" and "any" represent the options that were allowed as input and target lights. 
This any-to-any challenge was won by Puthussery \textit{et al.} \cite{puthussery2020wdrn}, who employed a wavelet transformation within their network. Wang \textit{et al.} \cite{wang2020deep} attained the highest PSNR in the challenge. This work \cite{wang2020deep} proposed a deep relighting network (DRN) comprising three parts: scene reconversion, shadow prior estimation, and rerendering.

Numerous subsequent articles have researched this topic \cite{kubiak2021silt, zhu2022ian, gafton20202d, dherse2020scene}. Specifically, Kubiak \textit{et al.} \cite{kubiak2021silt} proposed a self-supervised lighting transfer method capable of generating relit images with a uniform lighting style. They additionally incorporated intrinsic decomposition as extra guidance. However, these methods require the target illumination to be uniform and lack validation of their intrinsic decomposition output. Our dataset and method can bridge these gaps.
Zhu \textit{et al.} \cite{zhu2022ian} proposed the illumination-aware network (IAN), which leverages guidance from hierarchical sampling for high efficiency. While their method can handle various light inputs, it is only applied to "one-to-one" relighting on the Multi-Illumination and VIDIT datasets. In contrast, our work concentrates more on "any-to-any" relighting, which represents a more generalized issue.
\vspace{-7pt}
\subsection{Relighting from other perspectives}
We categorize relighting works from other perspectives into four distinct groups. The first group \cite{Dastjerdi_2023_ICCV, gardner2017learning, li2020inverse} concentrates on light estimation and inverse rendering, utilizing them as inputs for a rendering engine to relight an image or part of it. Dastjerdi \textit{et al.} \cite{Dastjerdi_2023_ICCV} proposed a method for editable lighting estimation by integrating a parametric light model with 360-degree panoramas, which is applicable to both indoor and outdoor environments.
The second group focuses on particular scenarios such as portrait scenes \cite{nestmeyer2020learning,sun2019single,zhou2019deep,pandey2021total, hou2022face}, or other specific objects \cite{sang2020single}. While these scenes are relatively fixed in terms of topology, the scenes we explore are considerably more complex. 
The third group \cite{duchene2015multi,xu2018deep,philip2019multi, moreno2005optimal, srinivasan2021nerv} addresses relighting via multiple images, which includes the work based on NeRF \cite{zhang2022invrender, NEURIPS2022_Hasselgren, srinivasan2021nerv, toschi2023relight}. Although NeRF can naturally manipulate the light conditions within a scene, a single trained model is only applicable to a single scene. 
The last group works on generative models such as StyleGAN \cite{karras2021alias}, aiming to create scenes under different illuminations \cite{forsyth2021sirfyn, bhattad2022enriching}. However, these networks are incapable of relighting a specified input image.
\vspace{-7pt}
\subsection{Intrinsic decomposition}
Since the seminal work of Barrow \textit{et al.} \cite{barrow1978recovering}, numerous methods have been proposed to address this problem \cite{barron2014shape, nestmeyer2017reflectance}. Most of the previous supervised approaches are based on deep architectures that extend the U-Net paradigm to a one-to-two encoder-decoder version \cite{Sial:20, baslamisli2018cnn, wang2019single, luo2020niid}. However, a major challenge for supervised intrinsic decomposition is the lack of sufficient GT data. To overcome this issue, several semisupervised and unsupervised methods have been introduced for estimating intrinsic decomposition \cite{li2018learning, lettry2018unsupervised, ma2018single, liu2020unsupervised, sengupta2019neural}. Among these methods, the method of Lettry \textit{et al.} \cite{lettry2018unsupervised} proposed a siamese training approach and introduced novel loss functions to capture intrinsic properties, thereby achieving unsupervised intrinsic decomposition. We use some of these ideas later in this work. 
\vspace{-14pt}
\section{New Datasets}

In this section, we introduce 2 new datasets specifically created for relighting. The first dataset is synthetic, whereas the second dataset is real. 

\vspace{-7pt}
\subsection{ISR: Intrinsic Scene Relighting Dataset}
The images of this first dataset are generated by the open-source Blender rendering engine following the methodology proposed by Sial et al. in \cite{Sial:20,sial2020light}. This dataset includes intrinsic components in the GT, thus it is called ISR for \textit{Intrinsic Scene Relighting}. 

The ISR has 7801 scenes, each one under 10 different light conditions. Each scene has between 3 and 10 nonoverlapping objects which are randomly selected from various categories of the \textit{ShapeNet} dataset \cite{Shi2017LearningNO} including electronics, pots, buses, cars, chairs, sofas, and airplanes. The blender {\it roughness} parameter is used to control the amount of light reflected from the object's surface.  We pay special attention to introducing background diversity and provoking complex lighting interactions in our scenes. The wall and floor in each scene are set differently, and the random choices include either homogeneous color or rich textured patterns from a Corel dataset \cite{Corel}. In addition, the walls have variable shapes, such as cylinders and polyhedrons. As a result, the dataset presents significant variations in reflectance and shading.

\begin{figure*}[htbp]
    \centering
    \begin{minipage}[b]{0.7\linewidth}
    \centering
    \includegraphics[width=\linewidth]{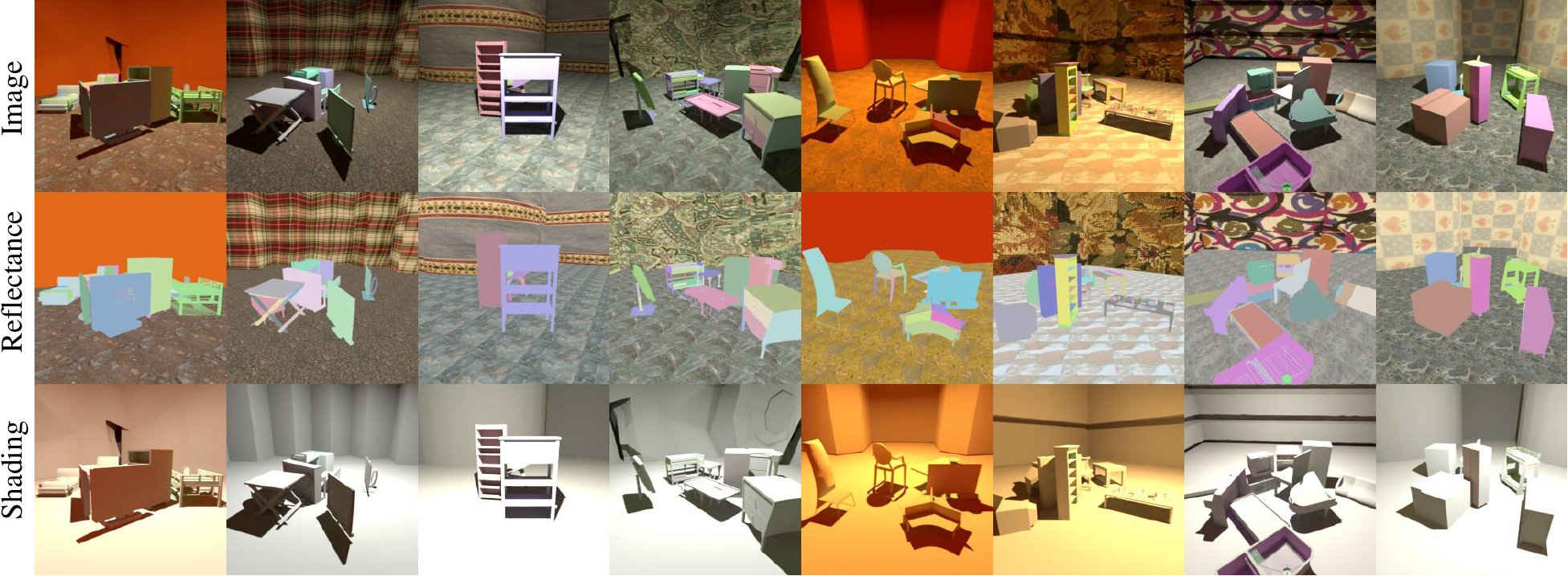}
    \small{(a)}
    \end{minipage}
    \begin{minipage}[b]{0.29\linewidth}
    \centering
    \includegraphics[width=\linewidth]{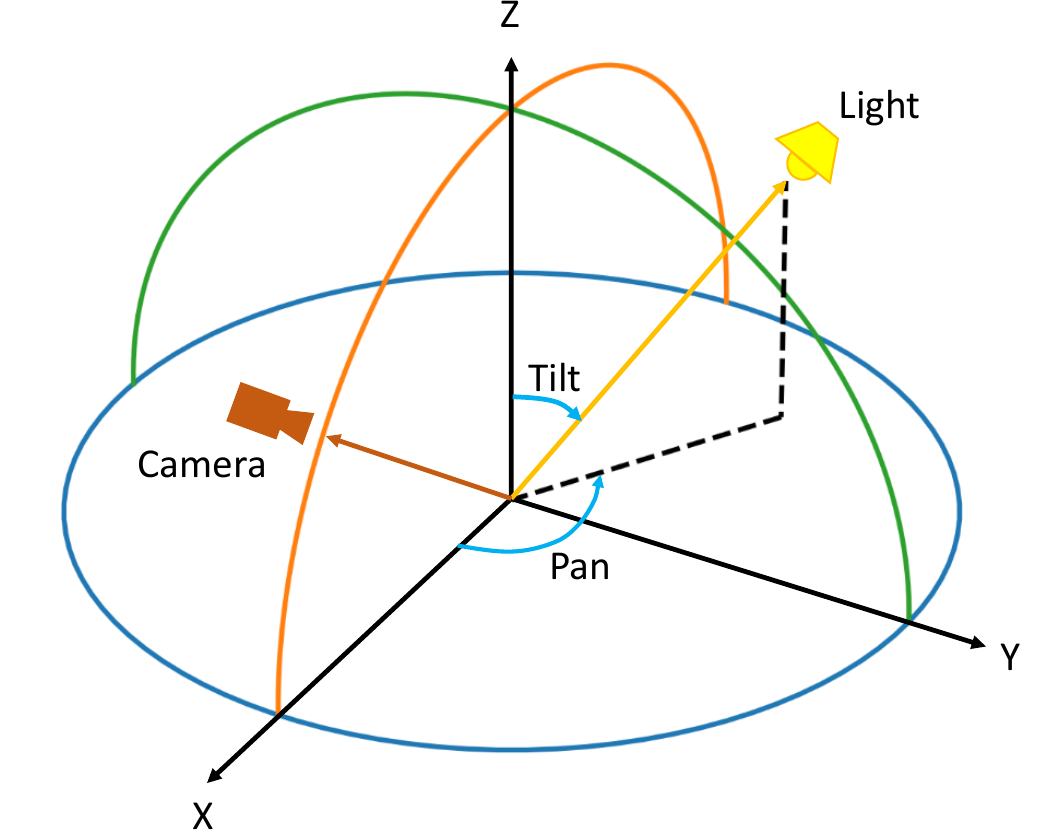}
    \small{(b)}
    \end{minipage}
    \centering
    \vspace{-20pt}
    \caption{ISR dataset. (a) Illustrations of some examples from the dataset, showing the images along with their corresponding reflectances and shading. (b) Coordinate system employed in the dataset, which includes the pan and tilt.}
    \label{fig:dataset}
    \vspace{-10pt}
\end{figure*}

Scenes are illuminated by a single oriented color source. We use Planckian lights, so the light color is represented by a single parameter, which is the color temperature. 
The light position is given by the pan and tilt angles, and is always oriented to the center of the scene. For each scene, 10 images are generated via a random position on an upper semisphere (see Fig.~\ref{fig:dataset}(a)) whose radius can randomly vary. Some image examples of the dataset are displayed in Fig.~\ref{fig:dataset}(a). The GT provides the 10 corresponding shading and reflectance components for each scene.
\vspace{-2pt}

\vspace{-8pt}
\subsection{RSR: Real scene relighting dataset}\label{subsec:RSR dataset}
\vspace{-2pt}
\begin{figure*}[htbp]
    \centering
    \begin{minipage}[c]{0.318\linewidth}
    \centering
    \includegraphics[width=\linewidth]{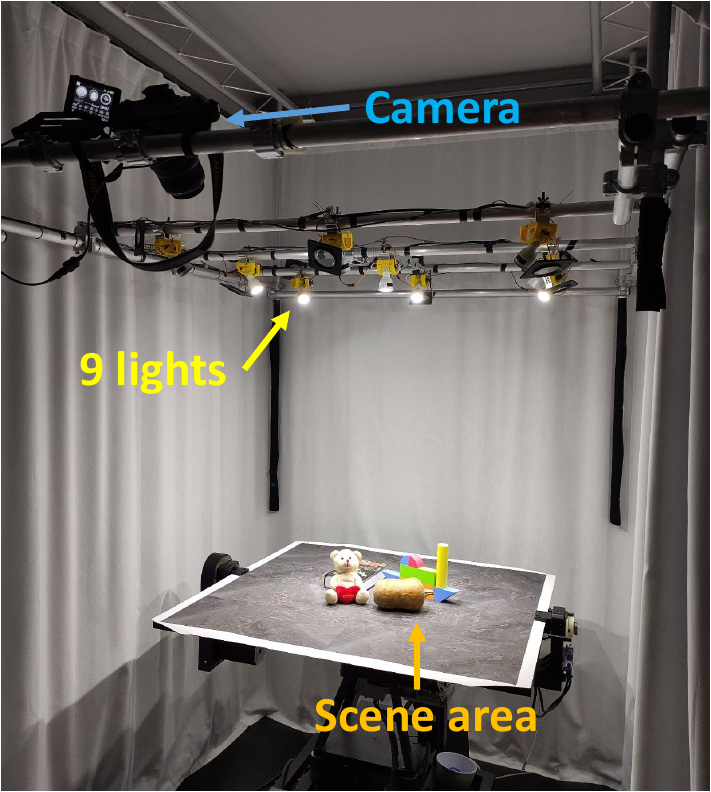}
    \small{(a)}
    \end{minipage}
    \begin{minipage}[c]{0.284\linewidth}
    \centering
    \includegraphics[width=\linewidth]{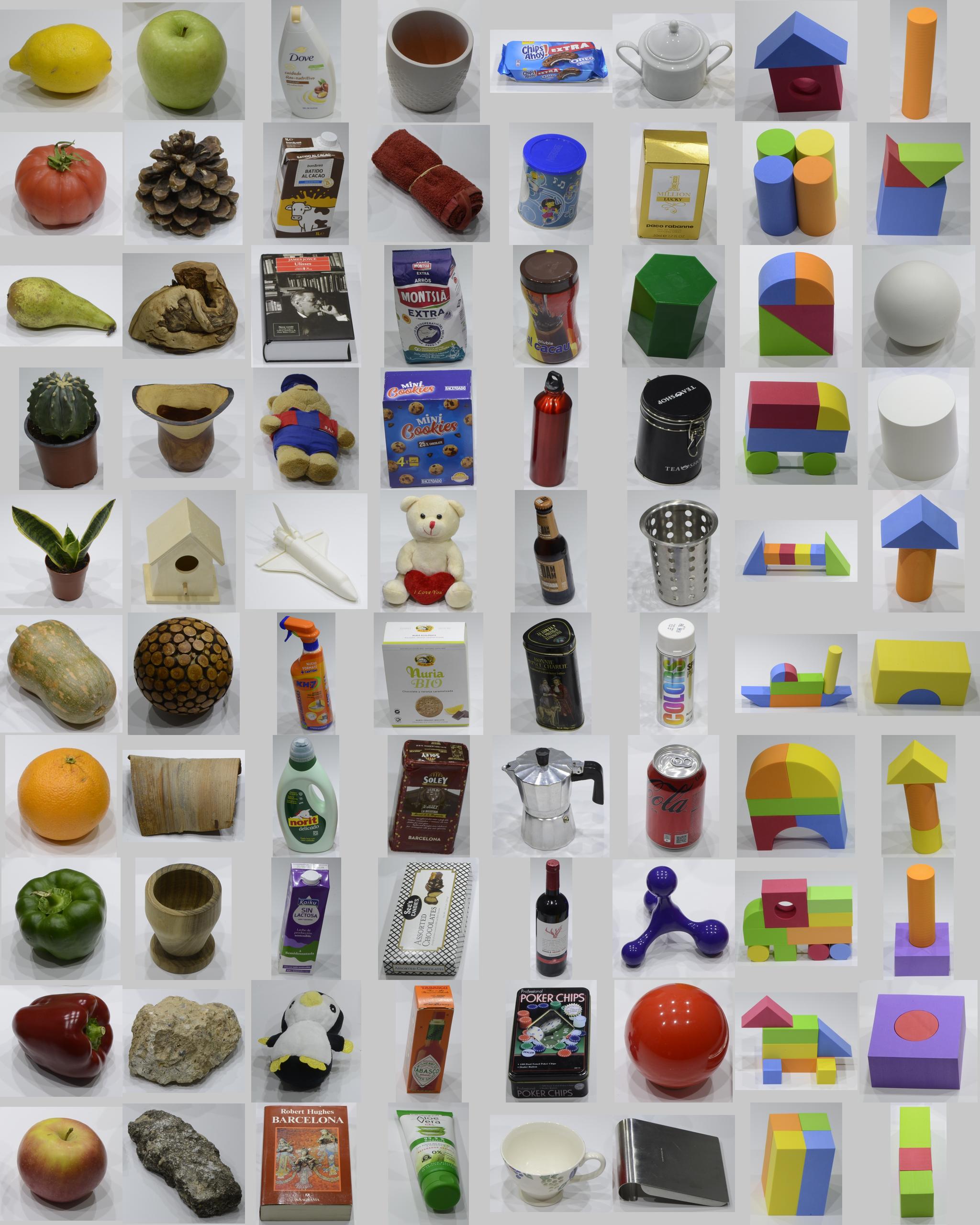}
    \small{(b)}
    \end{minipage}
    \begin{minipage}[c]{0.368\linewidth}
    \centering
    \includegraphics[width=\linewidth]{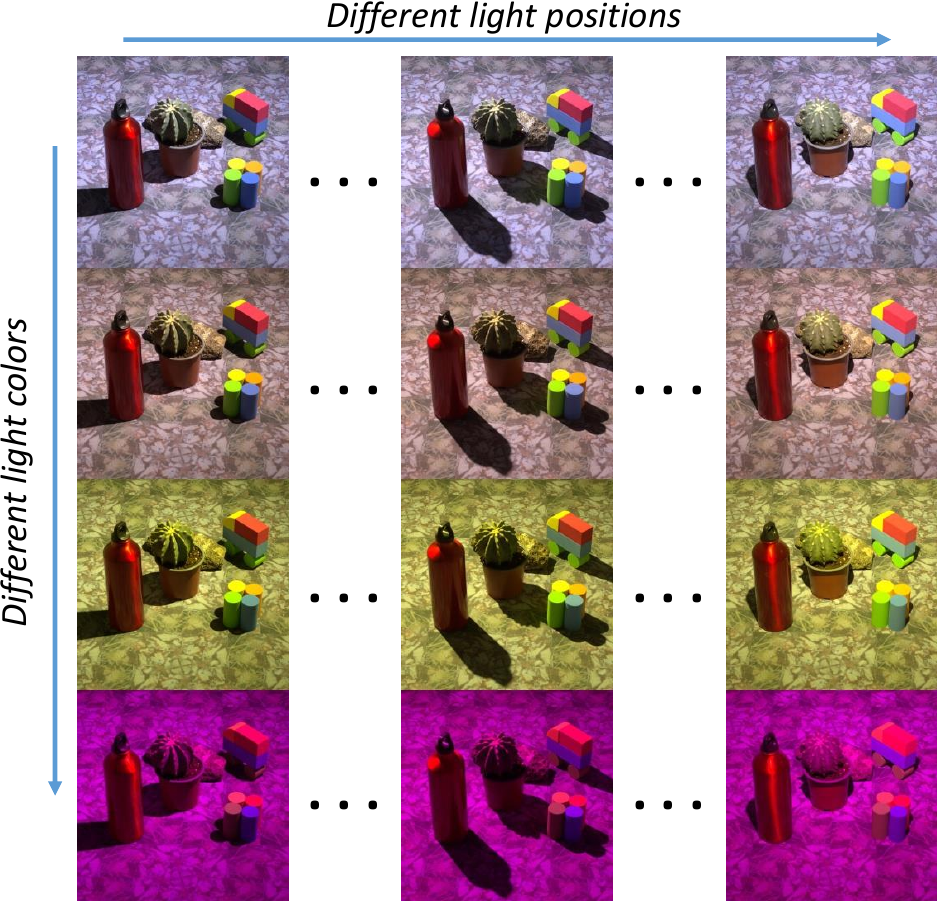}
    \small{(c)}
    \end{minipage}
    \centering
    \vspace{-3pt}
    \caption{RSR dataset. (a) Platform for building the RSR dataset. (b) The collection of objects in the dataset. (c) One configuration in the RSR dataset with different light positions and light colors.}
    \label{fig: rsr dataset}
    \vspace{-15pt}
\end{figure*}

To validate the effectiveness of our method in real-world scenes, we developed the \textit{real scene-relighting (RSR)} dataset. In synthetic environments, accurate simulations of the complex interactions of light with various objects can be challenging. Elements such as material properties and light behaviors are notably difficult to replicate in a computer-generated setting. Therefore, a real dataset, in addition to a synthetic dataset, is necessary. Furthermore, the ISR and RSR datasets share the same format for light representation, allowing us to validate the transferability of knowledge from the synthetic dataset to real-world scenes. Compared with the previous real dataset, the multi-illumination dataset \cite{murmann2019dataset}, our dataset features an explicit light representation, which enables us to evaluate our model under any target light conditions. The new dataset also includes numerous scenes with clear and complex cast shadows, enhancing the datasets available to the research community.

The RSR dataset is acquired in a lab with an automatic setup. We use a DSLR Nikon D5200 camera to capture the images. The scene area is lighted with a $3\times3$ multicolored led light matrix. The scene is set on a rotatable table. This acquisition system is shown in Fig.~\ref{fig: rsr dataset}(a). Each scene is composed of a background and several objects. The library of backgrounds has 8 textured options. 

We carefully choose a set of 80 real-life objects sampling a wide range of materials and shapes (see Fig.~\ref{fig: rsr dataset}(b)). From glossy to matte materials and from natural to basic shapes. The objects are organized into 3 groups, 20 natural (organic and stone), 40 manufactured (plastic, paper, glass, or metal), and 20 basic (foam) shapes.

To reduce human bias, scene configurations are randomly created by assigning indices to objects and backgrounds. Each configuration comprises 1 background and 2 to 5 objects. Each object is randomly positioned within the camera's field of view on the platform. Every configuration is captured with 8 distinct table rotations, which include 0\degree, $\pm$45\degree, $\pm$90\degree, $\pm$135\degree, and 180\degree, and under 36 varied light conditions, arising from a combination of 9 light positions and 4 distinct colors. The purpose of rotating the platform is to increase the number of scene points of view effectively. As a result, we obtain 72 unique configurations, which equal 576 scenes for relighting, for a total of 20,736 images.

\setlength{\tabcolsep}{1pt}
\begin{table}
\begin{center}
\vspace{-8pt}
\caption{Overview of the relighting datasets. For the VIDIT dataset, 300 of 390 scenes are fully released. }
\label{table: Overview of relighting datasets.}
\vspace{-5pt}
\begin{tabular}{ccccc}
\hline
Dataset            & Type      & Scene counts                 & Images/scene & Total images \\
\hline
Multi-illumination\cite{murmann2019dataset} & Real      & 1015                         & 25               & 25,375        \\
VIDIT\cite{helou2020vidit}              & Synthetic & 390(300) & 40               & 15,600(12,000)        \\
ISR                & Synthetic & 7801                         & 10               & 78,010        \\
RSR                & Real      & 576                          & 36               & 20,736       \\
\hline
\end{tabular}
\end{center}
\vspace{-22pt}
\end{table}
\setlength{\tabcolsep}{1pt}

Finally, Table \ref{table: Overview of relighting datasets.} provides an overview of both previous datasets and our newly proposed datasets. The size of our ISR dataset is highly competitive. Moreover, unlike those in the other three datasets, the light conditions in the ISR dataset are randomly sampled in a spatially continuous manner. Our RSR dataset is a real-world dataset, with a number of images close to that of the other real dataset, Multi-illumination. However, they differ in terms of scene characteristics. Additionally, the illumination representations also differ. The Multi-illumination dataset implicitly represents light conditions via probes, while the RSR dataset explicitly represents these conditions via positions and colors. Our two newly proposed datasets provide substantial supplements to research in this field.

\vspace{-8pt}
\section{Method}
\vspace{-3pt}
In this section, we explain our proposed deep architecture for single image relighting, and the light conditions can be any-to-any. We hypothesize that intrinsic decomposition is a robust physical constraint for the relighting problem, thus we design a two-stage architecture that follows this physics-guided approach. 
\vspace{-9pt}
\subsection{Relighting and Intrinsic Decomposition Constraint}
Intrinsic decomposition as proposed by Barrow and Tenenbaum \cite{barrow1978recovering} assumes that an image can be decomposed into the pixelwise product of two components, reflectance and shading:
\begin{equation}I(x, y)=R(x, y) \odot S(x, y)
\label{eq:Intrinsic decomposition}
\end{equation}
where $I$ is the resulting image, $R$ is the reflectance component, $S$ is the shading component, $I$, $R$, $S \in \mathbb{R} ^3$, and $(x,y)$ are pixel coordinates. This model assumes that the reflectance component is independent of the light condition, which affects only the shading component. We use this assumption as a physical constraint that forces the reflectance to be the same both for the input and relit images, while shading varies accordingly with the input and target light conditions. For the range, $I,R\in[0,1]^3$ and $S\in[0,+\infty]^3$. With this setting, $S$ can be colored and beyond unit value, which is also used in \cite{lettry2018unsupervised}. This model is more challenging than some traditional models, but it better adapts to the demands of relighting. For example, if shading can only be grayscale, it cannot account for the color of the light. If the range is constrained to 1, it cannot handle scenarios of specular reflection and overexposure. These scenarios are present in both previous relighting datasets \cite{helou2020vidit, murmann2019dataset}, and the datasets we propose.

\vspace{-9pt}
\subsection{Two-stage architecture for relighting}

\begin{figure*}
\centering
\includegraphics[width=0.95\linewidth]{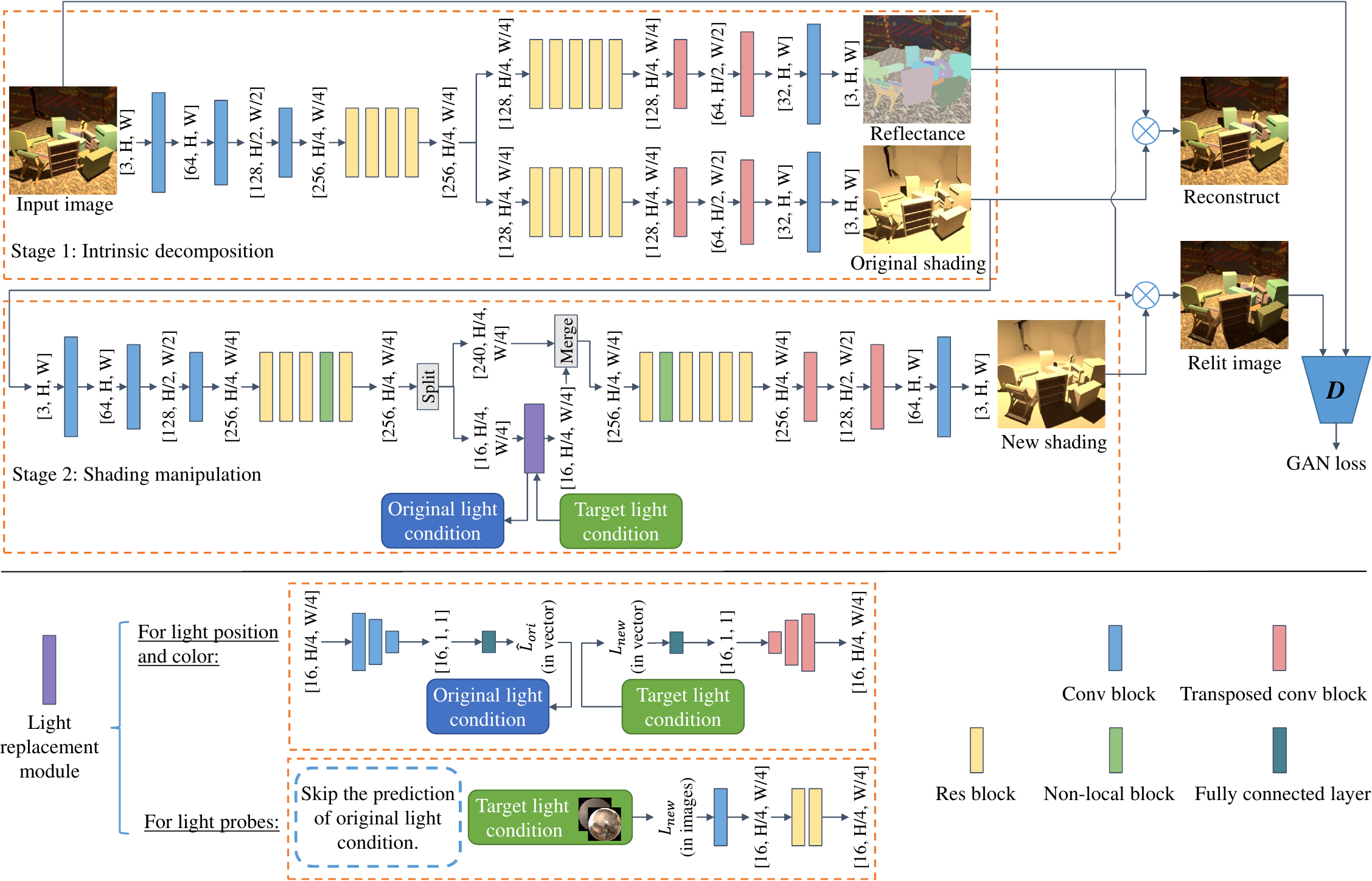}
\vspace{-4pt}
\caption{The architecture of our proposed two-stage network. Stage 1 is responsible for intrinsic decomposition, and Stage 2 carries out shading manipulation. Positioned in the middle of Stage 2 is a light replacement module, designed to predict the original light conditions and embed new ones. Owing to variations in light representation across different datasets, we develop distinct versions of the light replacement module to work on different datasets.}
\label{fig:architecture}
\vspace{-15pt}
\end{figure*}

We introduce a two-stage network architecture along with the constraint of intrinsic decomposition, which is shown in Fig.~\ref{fig:architecture}. Stage 1 carries out the intrinsic components, which are used in Stage 2 to generate the target shading.

Previous works \cite{pandey2021total, sun2019single, zhou2019deep} have mostly employed networks similar to U-Net\cite{ronneberger2015u}. While the original U-Net is capable of handling local features, its ability to capture long-range dependencies may be limited by the restricted receptive fields. However, long-range dependencies, such as cast shadows and their corresponding objects, are commonly encountered in relighting tasks. Consequently, we believe that the original U-Net backbone may not be sufficient for the relighting task. Therefore, we opt to use a ResNet backbone to replace the basic U-Net. In the subsequent ablation study, we also compare the results obtained via U-Net and ResNet backbone. Additionally, we incorporate non-local blocks in the network architecture to enhance the capture of long-range dependencies. 

In Stage 1, the input tensor $I_{input}(x,y) \in \mathbb{R}^{3}$, which is $h \times w$ large, is encoded through three convolution blocks, each consisting of Convolution-BatchNorm-ReLU layers. After each block, the resolution of the output is reduced, the channel depth is increased, and the final output tensor has dimensions $256 \times \frac{h}{4} \times \frac{w}{4}$.
Next, we use four residual blocks \cite{he2016deep} and split the output into two halves along channels. Each downstream branch then utilizes five additional residual blocks. Finally, we use two transposed convolution blocks and one convolution block to generate the output image. The upper branch predicts the reflectance component, whereas the lower branch predicts the original shading. Thus, Stage 1 can be formulated as follows:
\begin{equation}
(\hat{R}, \hat{S}_{ori}) = G_1(I_{input})
\label{eq: stage 1}
\end{equation}
where $\hat{R}$ and $\hat{S}_{ori}$ are the predictions of the reflectance and original shading of the input image, respectively. We hereby denote the predictions with hat, $\hat{X}$, and without hat for their GT counterparts.

In Stage 2, we also adopt a structure comprising an encoder, residual blocks, and a decoder, similar to Stage 1, but with several notable differences.
First, a light condition processing module is introduced in the middle of the architecture. The module splits the feature into two parts along the channels. The first part, with most of the channels, contains physical information about the scenes, whereas the second part, with fewer channels, contains information about the light condition. The details of this process are further described in Sec.~\ref{Light replacement module}.
Second, we utilize two non-local blocks. The first non-local block is positioned between the third and fourth residual blocks before the light condition is embedded. The second non-local block is placed between the first and second residual blocks after the light condition is embedded. Our motivation is that both the extraction of the light condition and the generation of the new shading rely on the long-range dependencies in the image feature. Overall, Stage 2 can be represented as:
\begin{equation}
(\hat{S}_{new}, \hat{\Phi}_{ori}) = G_2(\hat{S}_{ori}, \Phi_{new})
\label{eq: stage 2}
\end{equation}
where $\Phi_{new}$ is the input target light condition, and where $\hat{S}_{new}$ and $\hat{\Phi}_{ori}$ are the estimated target shading and the original light condition of the input image, respectively. Notably, the inclusion of the output $\hat{\Phi}_{ori}$ can be optional depending on the choice of the light replacement module. 

Once we obtain the target shading, we can create the relit image by multiplying the reflectance component from Stage 1 with the target shading from Stage 2:
\begin{equation}
I_{relit} = \hat{R} \odot \hat{S}_{new}
\label{eq: relit}
\end{equation}
Similarly, we can reconstruct the input image by multiplying the reflectance component with the original shading from Stage 1:
\begin{equation}
I_{recon} = \hat{R} \odot \hat{S}_{ori}
\label{eq: recon}
\end{equation}
Finally, we use a patch discriminator \cite{isola2017image} based on the LSGAN \cite{mao2017least} to enhance the perceptual quality of the relit image.
\vspace{-8pt}
\subsection{Light replacement module}\label{Light replacement module}
In the datasets used in this study, light conditions are represented in two forms. We design two networks to process them separately, as depicted at the bottom left of Fig.~\ref{fig:architecture}. The first network is designed for light conditions represented as a vector of position and color, as used in the ISR, RSR, and VIDIT datasets. The feature passes through several convolution layers and a fully connected layer to yield the light condition of the original image. The new light parameters are then fed into a fully connected layer and several transposed convolution layers. Finally, the shape of the output of the light replacement module is the same as its input. 

The second type is designed for light conditions represented as light probes, where the light information comes from the reflection on the spheres, as used in the Multi-Illumination dataset \cite{murmann2019dataset}. We use one convolution block and two residual blocks to transform the target light condition into a feature with the same shape as that of the previous type. However, we do not predict the original light condition because predicting the pictures of the probes is unnecessary and not feasible, as the mirror ball contains information outside the scene. 

\vspace{-8pt}
\subsection{Cross-relighting}
\begin{figure}[!t]
\centering
\includegraphics[width=\linewidth]{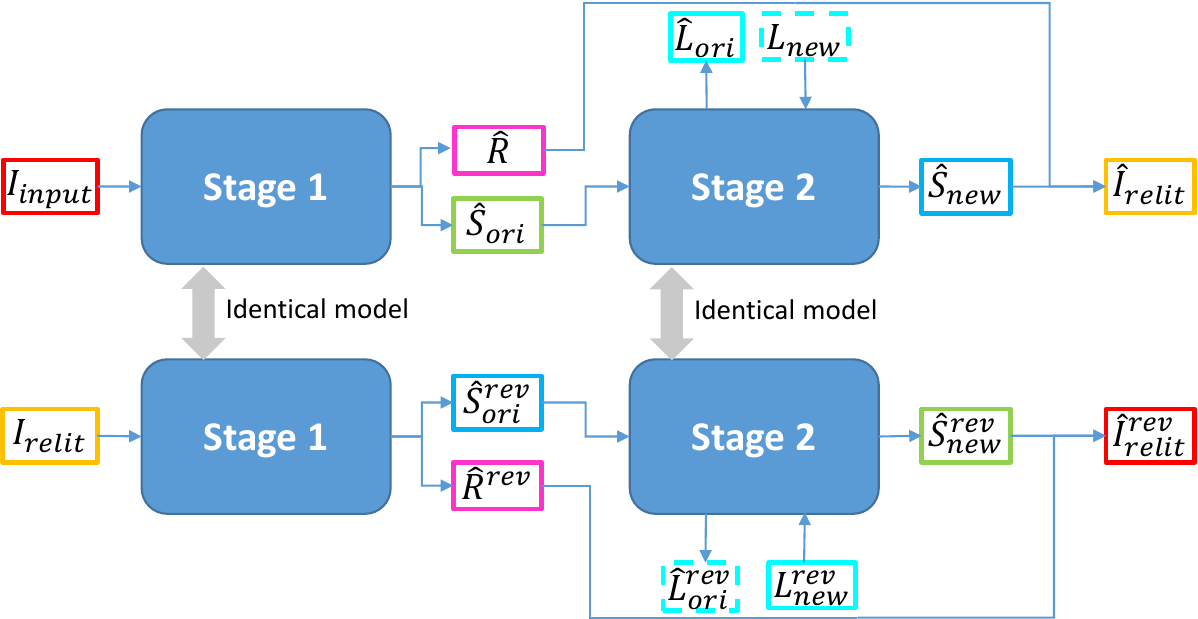}
\vspace{-14pt}
\caption{Diagram of cross-relighting. The variables of the reverse samples are denoted here with the superscript, $X^{rev}$. Boxes of the same color (or line type) indicate that the corresponding two elements should be the same or share the same GT.}
\vspace{-15pt}
\label{cross-relighting}
\end{figure}

Taking advantage of the reversibility deployed by the relighting process, we add a physical constraint to reinforce the training. The input and relit images can be interchanged through Stages 1 and 2, denoted as $S1+S2$, in this way:
\begin{equation}
I_{input} \; \xrightarrow[S1+S2]{\Phi_{new}} \; I_{relit}
\; , \;
I_{relit} \; \xrightarrow[S1+S2]{\Phi_{ori}}\;  \; I_{input}, 
\end{equation}
 Given the input image $I_{input}$ and the target light condition $\Phi_{new}$, the GT for the relit image is $I_{relit}$; conversely, if we input $I_{relit}$ and the $\Phi_{ori}$ (the light condition of $I_{input}$), the GT for the relit image is $I_{input}$. Similar findings can be found in \cite{zehni2021joint, kubiak2021silt}. We incorporate this constraint into the training process to facilitate better learning of the network. This constraint is named the 'cross-relighting' constraint, as illustrated in Fig.~\ref{cross-relighting}. In each training batch, the computations of the original batch samples are performed as shown in the upper part of the diagram. In addition, we create a reversed version as shown in the lower part. To improve efficiency, in actual implementation, we combine both versions along the batch dimension and perform forward computations together.

Cross-relighting not only serves as a means of data augmentation but also provides an opportunity to introduce some unsupervised constraints, including reflectance and shadow consistency. These constraints are further described in Sec. \ref{subsec:UIID}.

\vspace{-8pt}
\subsection{Supervised Losses}
When comparing the predicted image and the GT, we employ a combination of L1 loss, SSIM loss \cite{zhao2016loss}, and LPIPS loss \cite{zhang2018perceptual}. L1 and SSIM losses are categorized as distortion losses, whereas LPIPS is a perceptual loss. The aim is to pursue both accuracy and appearance simultaneously. It can be formulated as:
\begin{equation}
\mathcal{L}_{img}(\hat{x}, x) = \|\hat{x}-x\|_1 + (1 - SSIM(\hat{x}, x)) + LPIPS(\hat{x}, x)
\label{eq: loss}
\end{equation}
where $\hat{x}$ and $x$ are the predictions and the GT respectively. 
When the GT of intrinsic decomposition is available, we introduce losses for the final relit image and all internal predictions including the reflectance, original shading, new shading, and reconstruction of the original image from the predicted reflectance and original shading, as depicted in Fig.~\ref{fig:architecture}. 

For the light condition, the pan and tilt are constrained by angular loss, whereas the loss of light color is L1 loss directly:
\vspace{-5pt}
\begin{equation}
L_{light} = Angular(\hat{\Phi}_{ori, p}, \Phi_{ori, p}) + L1(\hat{\Phi}_{ori, c}, \Phi_{ori, c})
\label{eq: loss of light}
\end{equation}\label{eq:reverse}
where $\Phi_{ori, p}$ and $\Phi_{ori, c}$ are the light position and light color of $\Phi_{ori}$ respectively. 

The conditional GAN loss $L_{cGAN}$ and the losses of the discriminator use the same formulation proposed in \cite{mao2017least,isola2017image}, where all losses are summed as:
\vspace{-3pt}
\begin{equation}
\begin{split}
L_{all} &= \mathcal{L}_{img}(\hat{I}_{relit}, I_{relit}) + \mathcal{L}_{img}(\hat{I}_{recon}, I_{recon}) \\
&\quad + \mathcal{L}_{img}(\hat{R}, R) + \mathcal{L}_{img}(\hat{S}_{ori}, S_{ori}) \\
&\quad + \mathcal{L}_{img}(\hat{S}_{new}, S_{new}) 
+ L_{light} + L_{cGAN}
\end{split}
\label{eq: all loss}
\end{equation}
Finally, in cases where cross-relighting is employed for the reversed samples, this is denoted as $L_{cross}$, and it aligns with $L_{all}$. The final total loss can be expressed as follows:
\begin{equation}
L_{total} = L_{all} + L_{cross}
\label{eq: total loss}
\end{equation}

\subsection{Constraints for unsupervised intrinsic decomposition} \label{subsec:UIID}
\vspace{-2pt}
Introducing intrinsic decomposition is physically coherent and helpful, however, the GT of intrinsic decomposition is difficult to obtain in real datasets. Unsupervised learning has been frequently mentioned in research on intrinsic decomposition \cite{ma2018single, li2018learning, liu2020unsupervised, lettry2018unsupervised}. To handle the case where the GT for intrinsic decomposition is not available, our method incorporates an unsupervised component for intrinsic decomposition (abbreviated as UIID), which is inspired by \cite{lettry2018unsupervised}. This component introduces four unsupervised losses to disentangle the reflectance and shading. 

\paragraph{Reflectance Consistency}
As shown in Fig.~\ref{cross-relighting}, when we use cross-relighting, the network of stage 1 predicts the reflectance of one pair of images from the same scene, which by definition must be the same, this is what is called the reflectance consistency constraint:
\begin{equation}
L_{rc}=\|\hat{R}-\hat{R}^{rev}\|_1
\label{eq: Reflectance consistency}
\end{equation}
where $\hat{R}^{rev}$ denotes the estimated reflectance of $I_{relit}$, as shown in the bottom left of Fig.~\ref{cross-relighting}. The utilization of this operation is quite common in diverse unsupervised learning approaches for intrinsic decomposition. It has been employed in relighting research in combination with implicit intrinsic decomposition, as demonstrated by \cite{kubiak2021silt}. However, as suggested in \cite{lettry2018unsupervised}, solely relying on this operation leaves the problem underdetermined, resulting in training falling into the local pitfall. For example, the reflectance remains constant across all scenes, whereas only the shading varies with the input image. Therefore, additional constraints are necessary to address this limitation.

\paragraph{Shading Consistency}
Akin to reflectance, a similar constraint can be defined for shading. As illustrated in Fig.~\ref{cross-relighting}, owing to the reversibility of the cross-relighting, the shadings predicted in Stage 1, $\hat{S}_{ori}$ and $\hat{S}_{ori}^{rev}$, are expected to be equivalent to the shadings predicted in Stage 2, $\hat{S}_{new}^{rev}$ and $\hat{S}_{new}$, respectively. This concept can be referred to as shading consistency, which has been infrequently addressed in previous works. It can be introduced via the following loss:
\begin{equation}
L_{sc}=\|\hat{S}_{ori}-\hat{S}_{ori}^{rev}\|_1 + \|\hat{S}_{new}^{rev}-\hat{S}_{new}\|_1
\label{eq: Shading consistency}
\end{equation}

\paragraph{Shading Chromaticity Smoothness}
The luminance of shading can vary significantly with changes in object geometry, whereas the chromaticity of shading remains relatively stable because the light color in one scene is usually homogeneous. Thus, a shading chromaticity smoothness loss is introduced to constrain the gradient of the chromaticity of the shading. Previous work \cite{lettry2018unsupervised} constrained the chromaticity extracted from the ab dimensions of the CIE-Lab color space to 0. However, in the subsequent lines, we demonstrate that this constraint may be overly restrictive, as the gradient of shading chromaticity is nonzero. Therefore, we propose an alternative formulation for this constraint. 

In Table \ref{table: Statistics on the chromaticity}, we compute gradient statistics of images and intrinsic components on our ISR dataset. We list the average gradient of chromaticity calculated for all images, reflectances, and shadings. Chromaticity is computed on two different spaces after color space transformation from RGB: ab dimensions of CIE-Lab, and red-green and blue-yellow of the opponent space \cite{plataniotis2000color,sial2018color}. In this work, we opt to use the opponent space over the CIE-Lab, as it presents larger differences between reflectance and shading ($\nabla S/\nabla R$ according to Table \ref{table: Statistics on the chromaticity}). Additionally, it allows for more streamlined computations. Then, we calculate the gradient of the two chromaticity channels as well as all three channels. From our experiments, we find that the best way is to constrain $\nabla S/\nabla I$ to be lower than the mean value through an activation function such as the exponential linear unit (ELU)\cite{clevert2015fast}:
\begin{equation}f_{ac}(x)=\left\{\begin{array}{ll}
x + \alpha, & \text { if } x>0 \\
\alpha * \exp (x), & \text { if } x \leq 0
\end{array}\right.\end{equation}
where $\alpha =0.1$ in our experiments. The gradient of this function is 1 on the positive half-axis, whereas the gradient decreases significantly on the negative half-axis. The loss is formulated as:
\begin{equation}
\begin{aligned}
L_{scs} = &\lambda_1f_{ac}(\|\nabla{\hat{S}}^{opp,c}\|/\|\nabla I_{input}^{opp,c}\| - k_1) 
\\
& + \lambda_2f_{ac}(\|\nabla\hat{S}^{opp}\|/\|\nabla I_{input}^{opp}\| - k_2)
\end{aligned}
\label{eq: Shading Chromaticity Smooothness}
\end{equation}
where $\hat{S}^{opp}$ and $I_{input}^{opp}$ denote $\hat{S}$ and $I_{input}$ in the opponent space, respectively, while $\hat{S}^{opp,c}$ and $I_{input}^{opp,c}$ correspond to their respective chromaticity channels. The parameters we utilize include $k_1 = 0.5254$ and $k_2 = 0.7089$ (from Table \ref{table: Statistics on the chromaticity}), as well as the weights $\lambda_1=2.0$ and $\lambda_2=0.1$. We keep a small weight on the latter term involving all channels to avoid the luminance of the reflectance leaking to the shading. In addition, we use a customized $f_{ac}$ rather than other functions such as ReLU, to make the constraints less sensitive to the values of $k_1$ and $k_2$; this is facilitated by its gradient, which transitions smoothly at 0.

\setlength{\tabcolsep}{3pt}
\begin{table}
\begin{center}
\caption{Statistics of the gradient of chromaticity channels and all channels after the color space conversion on the ISR dataset}
\vspace{-4pt}
\label{table: Statistics on the chromaticity}
\begin{tabular}{ccccc}
\hline
\multicolumn{1}{l}{\multirow{2}{*}{}} & \multicolumn{2}{c}{CIE-Lab} & \multicolumn{2}{c}{Opponent color space} \\
\multicolumn{1}{l}{}                  & Chromaticity & All channels & Chromaticity        & All channels       \\ \hline
$\|\nabla I\|$                                     & 0.0058       & 0.0128       & 0.0118              & 0.0237             \\
$\|\nabla R\|$                                     & 0.0068       & 0.0139       & 0.0143              & 0.0275             \\
$\|\nabla S\|$                                     & 0.0031       & 0.0087       & 0.0062              & 0.0168             \\
$\|\nabla S\|/\|\nabla R\|$                                   & 0.4559	&0.6259	&0.4336	&0.6109 \\
$\|\nabla S\|/\|\nabla I\|$                                   & 0.5345	& 0.6797	& 0.5254	& 0.7089 \\
\hline
\end{tabular}
\end{center}
\vspace{-20pt}
\end{table}
\setlength{\tabcolsep}{3pt}

\paragraph{Initialization of Reflectance}
In the early stages of training, we introduce a loss for the initialization of reflectance, which is specifically designed to direct the color texture features of the original image toward reflectance, instead of shading. Since the input image and the relit image share the same reflectance, it would make sense to use either of them. \cite{lettry2018unsupervised} demonstrated that using a version with different illuminations from the input image would yield better results of intrinsic decomposition. Therefore, we encourage the reflectance to be close to the relit image:
\begin{equation}
L_{ir}=\mu(\|\hat{R}-I_{relit}\|_1 + \|\hat{R}^{rev}-I_{input}\|_1)
\label{eq: Initialization of Reflectance}
\end{equation}
where $\mu$ denotes a factor that initially decays from 1 to 0.01 during the first third of the training, and thereafter remains constant.

\paragraph{Total losses with UIID}
When training without the GT of intrinsic decomposition, the losses in Equ.\ref{eq: all loss} should be trimmed as: 
\begin{equation}
\begin{split}
L_{all}^* &= \mathcal{L}_{img}(\hat{I}_{relit}, I_{relit}) + \mathcal{L}_{img}(\hat{I}_{recon}, I_{recon}) \\
&\quad + L_{light} + L_{cGAN}
\end{split}
\label{eq: all loss without IID-GT}
\end{equation}
and the total loss with cross-relighting is formulated as:
\begin{equation}
\begin{split}
L_{total,u} &= L_{all}^* + L_{cross}^* + L_{rc} + L_{sc} 
\\
&\quad + L_{scs} + L_{scs}^* + L_{ir}
\end{split}
\label{eq: total loss without IID-GT}
\end{equation}
where $L_{cross}^*$ and $L_{scs}^*$ denote the corresponding losses for the reverse samples, paralleling $L_{cross}$ and $L_{scs}$ respectively. The remaining losses have already been derived based on cross-relighting.
\vspace{-6pt}
\section{Experiments and results}
\subsection{Implementation details and metrics}
We split the ISR dataset into training, validation, and test sets at a ratio of 85:5:10. Notably, when segmenting the RSR dataset, we utilize the configuration indices. This approach ensures that scenes from identical configurations are not distributed across distinct sets. Consequently, we divided the 72 configurations into 59 for training, 3 for validation, and 10 for testing.
Our network is implemented in PyTorch\cite{paszke2017automatic}. We use the Adam optimizer\cite{kingma2014adam}, a learning rate scheduler with step decay and a batch size of 18. The initial learning rate is set to 2e-4 for training on the ISR dataset, and 1e-4 for training or fine-tuning on other datasets. In all the experiments, we use an input and relit image resolution of 256$\times$256. For the VIDIT dataset, the images are resized, while for the Multi-illumination dataset, we first crop the sides because the aspect ratio is not equal to 1 before resizing.

To compare our method with other state-of-the-art (SOTA) methods, we train them on both our datasets and others. These methods include pix2pix\cite{isola2017image}, DRN\cite{wang2020deep} and IAN \cite{zhu2022ian}. Pix2pix is a typical image-to-image translation architecture consisting of a U-Net and a patch discriminator. The DRN achieves the best PSNR on track1 (one-to-one relighting) of the AIM2020 VIDIT challenge\cite{helou2020aim}. Their network contains two autoencoder networks with ResBlocks and a discriminator with multiscale perception. IAN is an illumination-aware network that achieves state-of-the-art (SOTA) performance in image relighting, and is capable of considering varying light conditions. However, the original Pix2pix and DRN can function only under one-to-one relighting conditions, meaning they do not consider light condition variations. Since our setting mainly focuses on different light conditions, we slightly modify the network and incorporate a module, akin to our method, which introduces light conditions in a suitable manner.

The metrics that we used to compare the prediction and the GT image include peak signal-to-noise ratio (PSNR), structural similarity index (SSIM) \cite{wang2004image}, learned perceptual image patch similarity (LPIPS) \cite{zhang2018unreasonable}, and mean perceptual score (MPS) \cite{helou2020aim}. 
The MPS \cite{helou2020aim} is the average of the SSIM and LPIPS scores, and is used as the final ranking metric in the AIM 2020 challenge \cite{helou2020aim}. It is defined as:
\begin{equation}
    MPS(x, y) = 0.5 \cdot (SSIM(x, y) + (1 - LPIPS(x, y)))
\end{equation}
where $SSIM(x, y)$ and $LPIPS(x, y)$ are the respective scores between image $x$ and image $y$.
\vspace{-8pt}
\subsection{Ablation study and results on our ISR dataset}

\setlength{\tabcolsep}{3pt}
\begin{table}
\begin{center}
\caption{Ablation study}
\label{table:ablation study}
\begin{tabular}{ccccc}
\hline
                                   & MPS $\uparrow$    & SSIM $\uparrow$   & LPIPS $\downarrow$  & PSNR $\uparrow$  \\ \hline
w/o ResNet(U-net instead) & 0.8908 & 0.8808 & 0.0993 & 23.40 \\ 
w/o non-local blocks               & 0.9172 & 0.9098 & 0.0754 & 25.09 \\
w/o two-stage model                & 0.9128 & 0.9042 & 0.0786 & 24.87 \\
w/o cross-relighting                    & 0.9191 & 0.9117 & 0.0735 & 25.44 \\
w/o lpips                          & 0.9147 & \textbf{0.9163} & 0.0868 & 25.72 \\
w/o ssim                           & 0.9177 & 0.9037 & \textbf{0.0684} & 25.59 \\ \hline
w/o IID-GT                      & 0.9152 & 0.9067 & 0.0763 & 25.06 \\
w/o IID-GT (w/ UIID) & 0.9178 & 0.9097 & 0.0740 & 25.34 \\ \hline
Full model                         & \textbf{0.9225} & 0.9151 & 0.0700 & \textbf{25.74} \\ \hline
\end{tabular}
\end{center}
\vspace{-18pt}
\end{table}
\setlength{\tabcolsep}{3pt}

\begin{figure*}[tb]
\centering
\hspace{-5pt}\begin{minipage}[b]{0.111\linewidth}
\centering
\small{Input}
\begin{tikzpicture}[spy using outlines={rectangle,yellow,magnification=2,size=\textwidth}]
\node {\includegraphics[width=\textwidth]{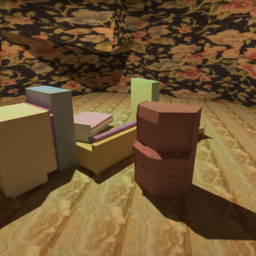}};
\spy on (0.1\textwidth,-0.1\textwidth) in node [left] at (0.5\textwidth,-\textwidth);
\end{tikzpicture}
\end{minipage}\hspace{-2pt}
\begin{minipage}[b]{0.111\linewidth}
\centering
\footnotesize{w/o ResNet}
\footnotesize{(U-net instead)}
\begin{tikzpicture}[spy using outlines={rectangle,yellow,magnification=2,size=\textwidth}]
\node {\includegraphics[width=\textwidth]{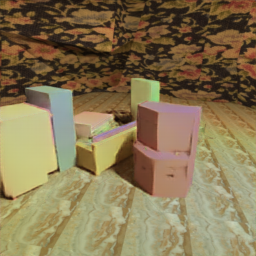}};
\spy on (0.1\textwidth,-0.1\textwidth) in node [left] at (0.5\textwidth,-\textwidth);
\end{tikzpicture}
\end{minipage}\hspace{-2pt}
\begin{minipage}[b]{0.111\linewidth}
\centering
\small{w/o non-local}
\begin{tikzpicture}[spy using outlines={rectangle,yellow,magnification=2,size=\textwidth}]
\node {\includegraphics[width=\textwidth]{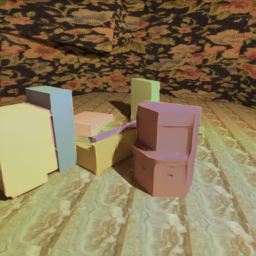}};
\spy on (0.1\textwidth,-0.1\textwidth) in node [left] at (0.5\textwidth,-\textwidth);
\end{tikzpicture}
\end{minipage}\hspace{-2pt}
\begin{minipage}[b]{0.111\linewidth}
\centering
\small{w/o two-stage}
\begin{tikzpicture}[spy using outlines={rectangle,yellow,magnification=2,size=\textwidth}]
\node {\includegraphics[width=\textwidth]{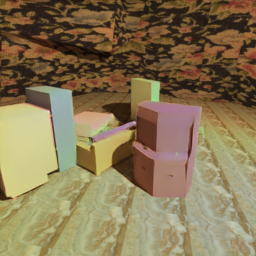}};
\spy on (0.1\textwidth,-0.1\textwidth) in node [left] at (0.5\textwidth,-\textwidth);
\end{tikzpicture}
\end{minipage}\hspace{-2pt}
\begin{minipage}[b]{0.111\linewidth}
\centering
\footnotesize{w/o cross-relighting}
\begin{tikzpicture}[spy using outlines={rectangle,yellow,magnification=2,size=\textwidth}]
\node {\includegraphics[width=\textwidth]{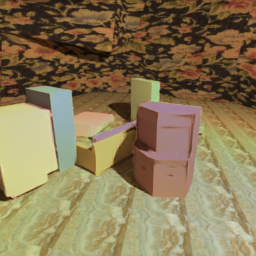}};
\spy on (0.1\textwidth,-0.1\textwidth) in node [left] at (0.5\textwidth,-\textwidth);
\end{tikzpicture}
\end{minipage}\hspace{-2pt}
\begin{minipage}[b]{0.111\linewidth}
\centering
\small{w/o lpips}
\begin{tikzpicture}[spy using outlines={rectangle,yellow,magnification=2,size=\textwidth}]
\node {\includegraphics[width=\textwidth]{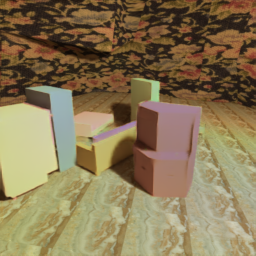}};
\spy on (0.1\textwidth,-0.1\textwidth) in node [left] at (0.5\textwidth,-\textwidth);
\end{tikzpicture}
\end{minipage}\hspace{-2pt}
\begin{minipage}[b]{0.111\linewidth}
\centering
\small{w/o ssim}
\begin{tikzpicture}[spy using outlines={rectangle,yellow,magnification=2,size=\textwidth}]
\node {\includegraphics[width=\textwidth]{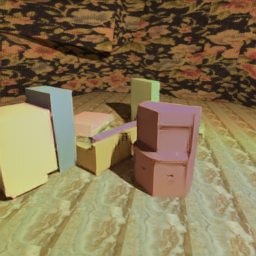}};
\spy on (0.1\textwidth,-0.1\textwidth) in node [left] at (0.5\textwidth,-\textwidth);
\end{tikzpicture}
\end{minipage}\hspace{-2pt}
\begin{minipage}[b]{0.111\linewidth}
\centering
\small{Full model}
\begin{tikzpicture}[spy using outlines={rectangle,yellow,magnification=2,size=\textwidth}]
\node {\includegraphics[width=\textwidth]{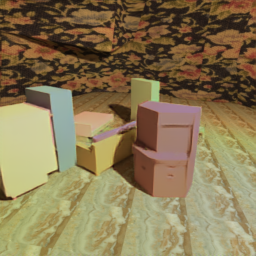}};
\spy on (0.1\textwidth,-0.1\textwidth) in node [left] at (0.5\textwidth,-\textwidth);
\end{tikzpicture}
\end{minipage}\hspace{-2pt}
\begin{minipage}[b]{0.111\linewidth}
\centering
\small{GT}
\begin{tikzpicture}[spy using outlines={rectangle,yellow,magnification=2,size=\textwidth}]
\node {\includegraphics[width=\textwidth]{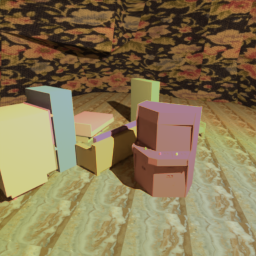}};
\spy on (0.1\textwidth,-0.1\textwidth) in node [left] at (0.5\textwidth,-\textwidth);
\end{tikzpicture}
\end{minipage}\hspace{-5pt}

\vspace{-2pt}

\hspace{-5pt}\begin{minipage}{0.111\linewidth}
\centering
\begin{tikzpicture}[spy using outlines={rectangle,yellow,magnification=2,size=\textwidth}]
\node {\includegraphics[width=\textwidth]{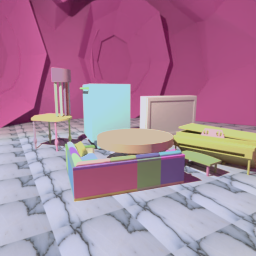}};
\spy on (-0.25\textwidth,-0.25\textwidth) in node [left] at (0.5\textwidth,-\textwidth);
\end{tikzpicture}
\end{minipage}\hspace{-2pt}
\begin{minipage}{0.111\linewidth}
\centering
\begin{tikzpicture}[spy using outlines={rectangle,yellow,magnification=2,size=\textwidth}]
\node {\includegraphics[width=\textwidth]{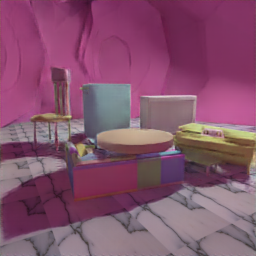}};
\spy on (-0.25\textwidth,-0.25\textwidth) in node [left] at (0.5\textwidth,-\textwidth);
\end{tikzpicture}
\end{minipage}\hspace{-2pt}
\begin{minipage}{0.111\linewidth}
\centering
\begin{tikzpicture}[spy using outlines={rectangle,yellow,magnification=2,size=\textwidth}]
\node {\includegraphics[width=\textwidth]{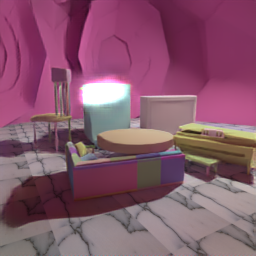}};
\spy on (-0.25\textwidth,-0.25\textwidth) in node [left] at (0.5\textwidth,-\textwidth);
\end{tikzpicture}
\end{minipage}\hspace{-2pt}
\begin{minipage}{0.111\linewidth}
\centering
\begin{tikzpicture}[spy using outlines={rectangle,yellow,magnification=2,size=\textwidth}]
\node {\includegraphics[width=\textwidth]{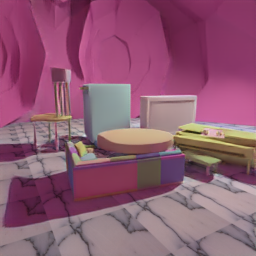}};
\spy on (-0.25\textwidth,-0.25\textwidth) in node [left] at (0.5\textwidth,-\textwidth);
\end{tikzpicture}
\end{minipage}\hspace{-2pt}
\begin{minipage}{0.111\linewidth}
\centering
\begin{tikzpicture}[spy using outlines={rectangle,yellow,magnification=2,size=\textwidth}]
\node {\includegraphics[width=\textwidth]{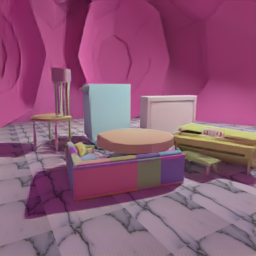}};
\spy on (-0.25\textwidth,-0.25\textwidth) in node [left] at (0.5\textwidth,-\textwidth);
\end{tikzpicture}
\end{minipage}\hspace{-2pt}
\begin{minipage}{0.111\linewidth}
\centering
\begin{tikzpicture}[spy using outlines={rectangle,yellow,magnification=2,size=\textwidth}]
\node {\includegraphics[width=\textwidth]{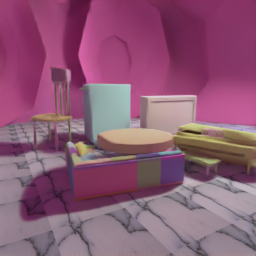}};
\spy on (-0.25\textwidth,-0.25\textwidth) in node [left] at (0.5\textwidth,-\textwidth);
\end{tikzpicture}
\end{minipage}\hspace{-2pt}
\begin{minipage}{0.111\linewidth}
\centering
\begin{tikzpicture}[spy using outlines={rectangle,yellow,magnification=2,size=\textwidth}]
\node {\includegraphics[width=\textwidth]{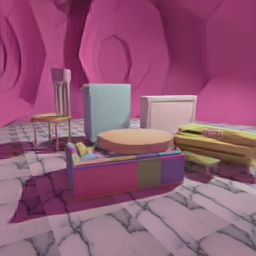}};
\spy on (-0.25\textwidth,-0.25\textwidth) in node [left] at (0.5\textwidth,-\textwidth);
\end{tikzpicture}
\end{minipage}\hspace{-2pt}
\begin{minipage}{0.111\linewidth}
\centering
\begin{tikzpicture}[spy using outlines={rectangle,yellow,magnification=2,size=\textwidth}]
\node {\includegraphics[width=\textwidth]{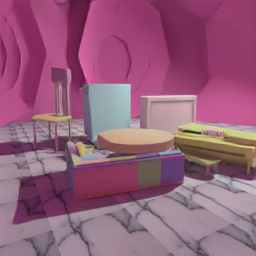}};
\spy on (-0.25\textwidth,-0.25\textwidth) in node [left] at (0.5\textwidth,-\textwidth);
\end{tikzpicture}
\end{minipage}\hspace{-2pt}
\begin{minipage}{0.111\linewidth}
\centering
\begin{tikzpicture}[spy using outlines={rectangle,yellow,magnification=2,size=\textwidth}]
\node {\includegraphics[width=\textwidth]{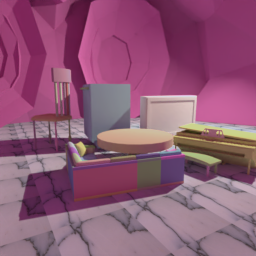}};
\spy on (-0.25\textwidth,-0.25\textwidth) in node [left] at (0.5\textwidth,-\textwidth);
\end{tikzpicture}
\end{minipage}\hspace{-5pt}
\caption{Qualitative results of the ablation study with access to the IID-GT on the ISR dataset.}
\vspace{-15pt}
\label{fig: Qualitative results of ablation study on the ISR}
\end{figure*}

\begin{figure*}[htbp]
   \begin{minipage}[c]{0.84\textwidth}
    \includegraphics[width=1.0\linewidth]{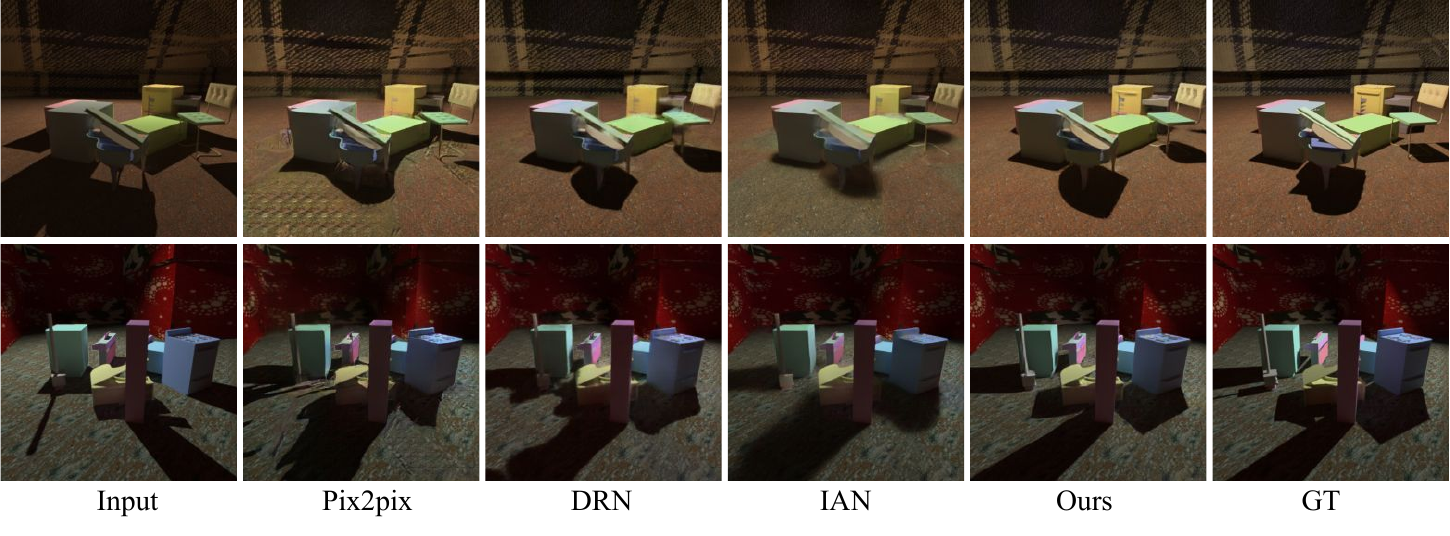}
  \end{minipage}\hfill
  \begin{minipage}[c]{0.15\textwidth}
    \caption{Qualitative comparison with other methods on the ISR dataset.} \label{fig: Qualitative comparison with other methods on the ISR dataset.}
  \end{minipage}
   \vspace{-15pt}
\end{figure*}

In this section, we conduct an ablation study on the various components of our proposed network using our ISR dataset. Initially, we analyze the case with access to the GT of intrinsic decomposition (abbreviated as IID-GT). The quantitative and qualitative results are shown in Table \ref{table:ablation study} and Fig.~\ref{fig: Qualitative results of ablation study on the ISR}. Subsequently, we investigate the performance in the absence of the IID-GT, and for this case, we analyze the effects before and after incorporating UIID constraints, and examine the role of losses in the UIID. The results are presented in Table \ref{table:Analysis of UIID}, Fig.~\ref{fig: Study for w/o IID-GT, w/o IID-GT (w/ UIID), and Full Model: Qualitative results of intrinsic decomposition and relighting on the ISR dataset.} and Fig. \ref{fig: Ablation study for the UIID module: Qualitative results of intrinsic decomposition and relighting on the ISR dataset.}. Finally, we train and evaluate previous methods to perform a comparative analysis with our approach. The results are illustrated in Table \ref{table:Compare with other methods on ISR dataset} and Fig.~\ref{fig: Qualitative comparison with other methods on the ISR dataset.}.

\paragraph{Ablation study with intrinsic GT}

In Table \ref{table:ablation study} and Fig.~\ref{fig: Qualitative results of ablation study on the ISR}, we initially conducted a comparison between the U-net and ResNet backbone in the row "w/o ResNet(U-net instead)". In this experiment, the non-local block was not utilized due to implementation incompatibility. The results clearly indicate that the ResNet backbone significantly outperforms U-Net. Meanwhile, the results in Fig.~\ref{fig: Qualitative results of ablation study on the ISR} demonstrate that the output of the Full model significantly surpasses that of the U-net in terms of appearance, and is much closer to the GT.

The term "w/o non-local blocks" indicates that we removed non-local blocks in Stage 2 of the network. Similarly, "w/o two-stage model" means the exclusion of the intrinsic decomposition part (Stage 1), performing direct relighting using only Stage 2. "w/o cross-relighting" refers to disabling the training process based on reverse samples and discarding $L_{cross}$ in Equation \ref{eq: total loss}. The table clearly shows that removing any of the three components leads to a decrease in the quantitative results across all the metrics. Additionally, as demonstrated in Fig.~\ref{fig: Qualitative results of ablation study on the ISR}, their results, compared with those of the full model, display peculiar cast shadow artifacts and a decline in image quality.

For the study of losses, "w/o lpips" and "w/o ssim" indicate the removal of LPIPS and SSIM loss from Equ. \ref{eq: loss} in the full model, respectively. Compared with the full model, removing LPIPS loss leads to an increase in the SSIM metric, whereas removing SSIM loss results in an improvement in the LPIPS metric. However, as shown in Fig.~\ref{fig: Qualitative comparison with other methods on the ISR dataset.}, in the results of "w/o lpips", the edges of the shadows become blurry. In the results of "w/o ssim", the shape of the shadows deviates significantly from the GT. Therefore, to achieve better visual appearance, we believe that it is necessary to retain both LPIPS and SSIM losses in the full model.

\begin{figure}
    \centering
    \begin{minipage}[c]{0.4\linewidth}
    \raggedright
    \hspace{-8pt}\rotatebox{90}{\centering\small{\quad \quad Input}}     \includegraphics[width=0.575\linewidth]{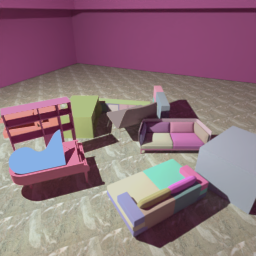}
    \end{minipage}
    \hspace{-10pt}\begin{minipage}[c]{0.6\linewidth}
    \caption{Study for w/o IID-GT, w/o IID-GT (w/ UIID), and Full model: Qualitative results of intrinsic decomposition and relighting on the ISR dataset.}
    \label{fig: Study for w/o IID-GT, w/o IID-GT (w/ UIID), and Full Model: Qualitative results of intrinsic decomposition and relighting on the ISR dataset.}
    \end{minipage}
    \begin{minipage}[b]{\linewidth}
    \vspace{3pt}
    \centering
    \setlength{\tabcolsep}{1pt}
    \begin{tabular}{ccccc}
    \hspace{-3pt}\rotatebox{90}{\small{Reflectance}} & \includegraphics[width=0.23\linewidth]{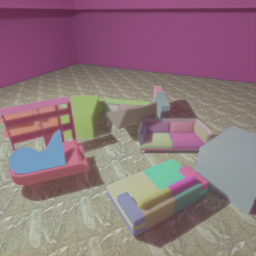} & \includegraphics[width=0.23\linewidth]{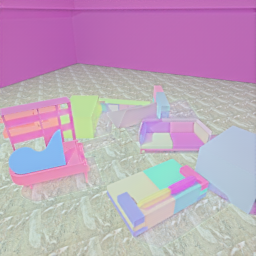} & \includegraphics[width=0.23\linewidth]{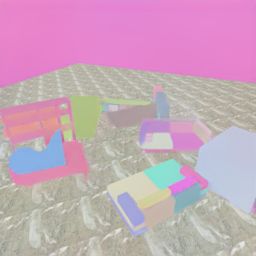} & \includegraphics[width=0.23\linewidth]{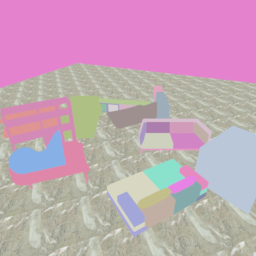} \\
    \hspace{-3pt}\rotatebox{90}{\footnotesize{Original Shading}} & \includegraphics[width=0.23\linewidth]{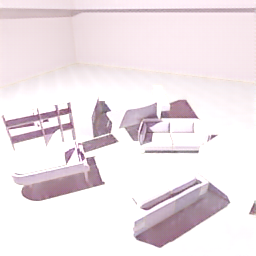} & \includegraphics[width=0.23\linewidth]{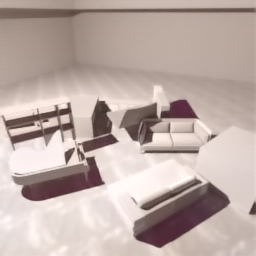} & \includegraphics[width=0.23\linewidth]{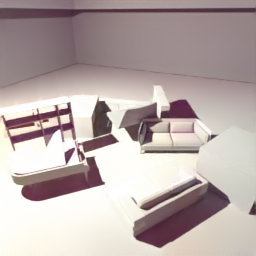} & \includegraphics[width=0.23\linewidth]{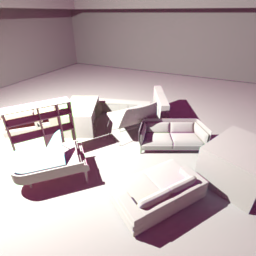} \\
    \hspace{-3pt}\rotatebox{90}{\small{New Shading}} & \includegraphics[width=0.23\linewidth]{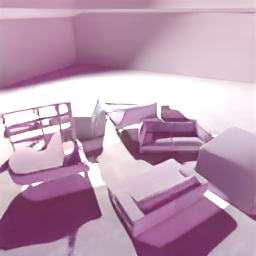} & \includegraphics[width=0.23\linewidth]{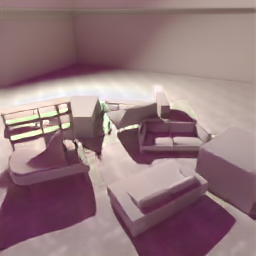} & \includegraphics[width=0.23\linewidth]{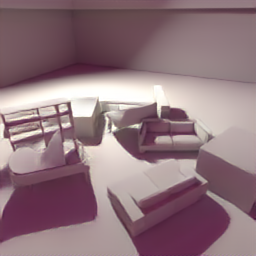} & \includegraphics[width=0.23\linewidth]{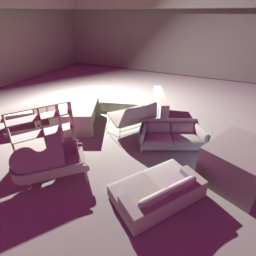} \\
    \hspace{-3pt}\rotatebox{90}{\small{\quad \quad Relit}} & \includegraphics[width=0.23\linewidth]{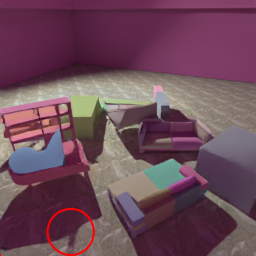} & \includegraphics[width=0.23\linewidth]{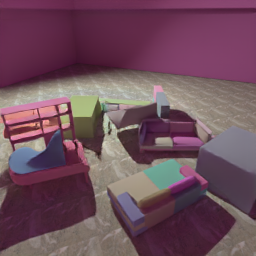} & \includegraphics[width=0.23\linewidth]{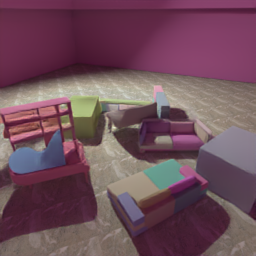} & \includegraphics[width=0.23\linewidth]{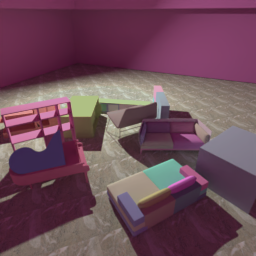} \\
    \vspace{-10pt}
     & \multirow{2}{*}{\small{w/o IID-GT}} & \multirow{2}{*}{\shortstack{\small{w/o IID-GT}\\\small{(w/ UIID)}}} & \multirow{2}{*}{\small{Full model}} & \multirow{2}{*}{\small{GT}}
    \end{tabular}
    \setlength{\tabcolsep}{1pt}
    
    \end{minipage}
\end{figure}

\paragraph{Study without intrinsic GT}
The case of "w/o IID-GT" involves training the network without the availability of the IID-GT. This is achieved by directly removing the losses $\mathcal{L}_{img}(\hat{R}, R)$, $\mathcal{L}_{img}(\hat{S}_{ori}, S_{ori})$, and $\mathcal{L}_{img}(\hat{S}_{new}, S_{new})$ from Equ. \ref{eq: all loss}. As shown in Table \ref{table:ablation study}, the quantitative results noticeably decrease compared with those of the full model, confirming our hypothesis that intrinsic disentangling helps in the relighting task.

However, the lack of IID-GT can be compensated for by incorporating the proposed  UIID component, as discussed in Sec.~\ref{subsec:UIID}, and the quantitative results improve and approach those of the full model. Moreover, Fig.~\ref{fig: Study for w/o IID-GT, w/o IID-GT (w/ UIID), and Full Model: Qualitative results of intrinsic decomposition and relighting on the ISR dataset.} displays the qualitative results of the aforementioned two experiments and the full model (trained with the IID-GT). We can see that the full model achieves satisfactory results in terms of relit images, reflectance, and shading because they are all supervised. When the constraints on the intrinsic components are removed ("w/o IID-GT" experiment), noticeable errors emerge in both reflectance and original shading. The new shading still retains traces of the previously cast shadows, and anomalous shadows emerge (indicated with a red circle). After incorporating UIID, the results of intrinsic decomposition become remarkably closer to those of the full model. Although there are still some ground texture artifacts present in the shading, the overall outcome is quite acceptable, and the final relighting results are also improved.

Furthermore, we investigate the role of each loss in the UIID component. Table \ref{table:Analysis of UIID} presents the quantitative results of the ablation study pertaining to the four loss functions in the UIID component, and the symbols for each term can be found in Equation \ref{eq: total loss without IID-GT}. These results encompass not only the relit images but also the intermediate outputs related to intrinsic decomposition. As shown in Table \ref{table:Analysis of UIID}, the removal of any one of the losses leads to a significant deterioration in the intrinsic decomposition results, including the reflectance, the original shading, and the new shading. Moreover, the elimination of these losses causes minor fluctuations in the relighting results. Although the exclusion of certain losses might slightly increase the relighting results, we still retain all the loss functions to avoid a substantial decline in the performance of intrinsic decomposition. 

Figure \ref{fig: Ablation study for the UIID module: Qualitative results of intrinsic decomposition and relighting on the ISR dataset.} illustrates the qualitative results of this ablation study. When the reflectance consistency and the shading consistency are removed, the predicted reflectance erroneously retains remnants of cast shadows, whereas concurrently, the new shading has a distinct effect on double-cast shadows (indicated with red circles). Additionally, the impact of omitting the shading consistency is more pronounced than that of removing the reflectance consistency. If the shading chromaticity smoothness $L_{scs}+L_{scs}^*$ is omitted, it results in a noticeable leakage of object colors into the shading, whereas these colors should ideally be present only in the reflectance. If we remove the initialization of the reflectance $L_{ir}$, the color texture of the background incorrectly seeps into the original shading to a greater extent (indicated with a red circle). While this error is not entirely mitigated in the results of the full UIID, it is still significantly reduced. Moreover, the relit image with the full UIID more closely resembles the GT.

\setlength{\tabcolsep}{3pt}
\begin{table}
\begin{center}
\caption{Analysis of UIID}
\label{table:Analysis of UIID}
\begin{tabular}{cccccc}
\hline
                              &          & MPS $\uparrow$    & SSIM $\uparrow$   & LPIPS $\downarrow$  & PSNR $\uparrow$  \\ \hline
\multirow{5}{*}{Reflectance}  
                              & w/o $L_{rc}$ & 0.8016 & 0.8252 & 0.2220 & 16.86 \\
                              & w/o $L_{sc}$ & 0.7499 & 0.7571 & 0.2573 & 13.06 \\
                              & w/o $L_{scs} + L_{scs}^*$ & 0.7869 & 0.8120 & 0.2381 & 15.00 \\
                              & w/o $L_{ir}$ & 0.7565 & 0.7947 & 0.2817 & 14.86 \\
                              & w/ full UIID     & \textbf{0.8270} & \textbf{0.8536} & \textbf{0.1997} & \textbf{18.15} \\
\hline
\multirow{5}{*}{\begin{tabular}[c]{@{}c@{}}Original \\ shading\end{tabular}} 
                              & w/o $L_{rc}$ & 0.7518 & 0.8376 & 0.3340 & 16.27 \\
                              & w/o $L_{sc}$ & 0.7129 & 0.7864 & 0.3607 & 12.13 \\
                              & w/o $L_{scs} + L_{scs}^*$ & 0.7511 & 0.8036 & \textbf{0.3015} & 15.06 \\
                              & w/o $L_{ir}$ & 0.7069 & 0.7807 & 0.3668 & 14.95 \\
                              & w/ full UIID     & \textbf{0.7691} & \textbf{0.8515} & 0.3133 & \textbf{17.30} \\
\hline
\multirow{5}{*}{\begin{tabular}[c]{@{}c@{}}New \\ shading\end{tabular}} 
                              & w/o $L_{rc}$ & 0.7011 & 0.7586 & 0.3563 & 14.92 \\
                              & w/o $L_{sc}$ & 0.6725 & 0.7191 & 0.3742 & 11.61 \\
                              & w/o $L_{scs} + L_{scs}^*$ & 0.6952 & 0.7365 & 0.3462 & 14.10 \\
                              & w/o $L_{ir}$ & 0.6582 & 0.7084 & 0.3919 & 13.92 \\
                              & w/ full UIID     & \textbf{0.7144} & \textbf{0.7714} & \textbf{0.3427} & \textbf{15.57} \\
\hline
\multirow{5}{*}{\begin{tabular}[c]{@{}c@{}}Relit \\ image\end{tabular}}        
                              & w/o $L_{rc}$ & 0.9181 & 0.9102 & 0.0740 & 25.46 \\
                              & w/o $L_{sc}$ & 0.9148 & 0.9067 & 0.0772 & 25.21 \\
                              & w/o $L_{scs} + L_{scs}^*$ & \textbf{0.9211} & \textbf{0.9135} & \textbf{0.0713} & \textbf{25.73} \\
                              & w/o $L_{ir}$ & 0.9189 & 0.9110 & 0.0732 & 25.49 \\
                              & w/ full UIID     & 0.9178 & 0.9097 & 0.0740 & 25.34 \\
\hline
\end{tabular}
\end{center}
\vspace{-10pt}
\end{table}
\setlength{\tabcolsep}{3pt}

\newlength{\imguiidsize}
\setlength{\imguiidsize}{0.25\linewidth}
\begin{figure*}
    \centering
\begin{minipage}[c]{0.14\linewidth}
\centering
    \hspace{-8pt}
    \rotatebox{90}{\centering\small{\quad \quad Input}}     \includegraphics[width=\imguiidsize]{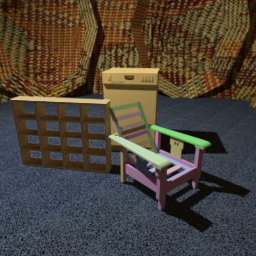}
    \hspace{-10pt}
    \caption{Ablation study for the UIID module: Qualitative results of intrinsic decomposition and relighting on the ISR dataset.}
    \label{fig: Ablation study for the UIID module: Qualitative results of intrinsic decomposition and relighting on the ISR dataset.}
\centering
\end{minipage}
\begin{minipage}[b]{0.84\linewidth}
    \vspace{3pt}
    \centering
    \setlength{\tabcolsep}{1pt}
\begin{tabular}{ccccccc}
\hspace{-3pt}\rotatebox{90}{\small{\quad Reflectance}} & \includegraphics[width=\imguiidsize]{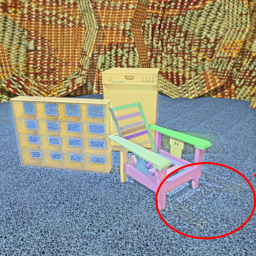} & \includegraphics[width=\imguiidsize]{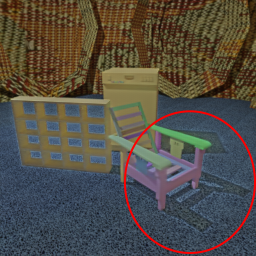} & \includegraphics[width=\imguiidsize]{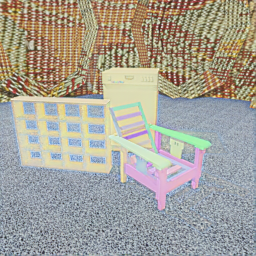} & \includegraphics[width=\imguiidsize]{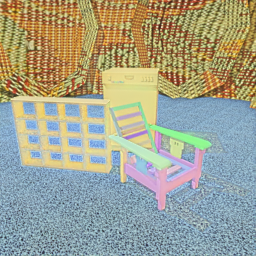} & \includegraphics[width=\imguiidsize]{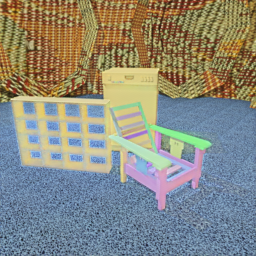} & \includegraphics[width=\imguiidsize]{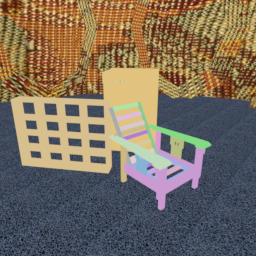} \\
\hspace{-3pt}\rotatebox{90}{\footnotesize{Original Shading}} & \includegraphics[width=\imguiidsize]{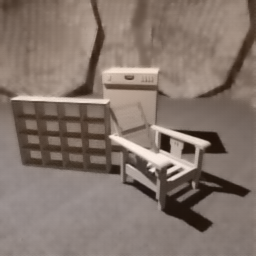} & \includegraphics[width=\imguiidsize]{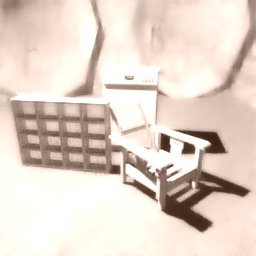} & \includegraphics[width=\imguiidsize]{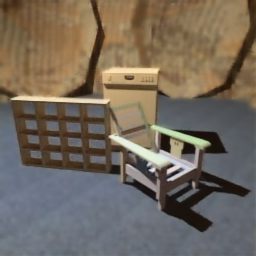} & \includegraphics[width=\imguiidsize]{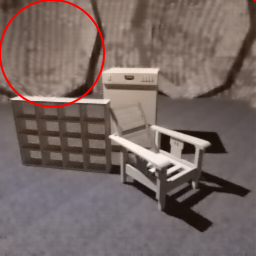} & \includegraphics[width=\imguiidsize]{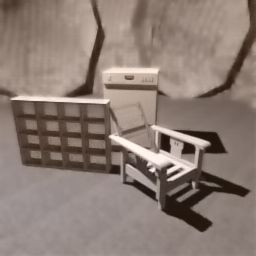} & \includegraphics[width=\imguiidsize]{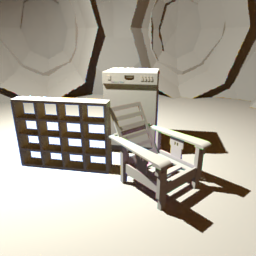} \\
\hspace{-3pt}\rotatebox{90}{\small{    New Shading}} & \includegraphics[width=\imguiidsize]{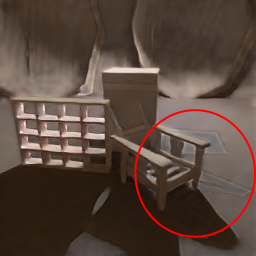} & \includegraphics[width=\imguiidsize]{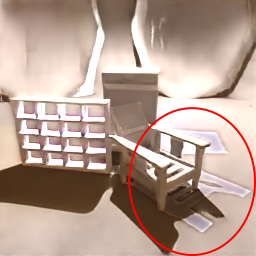} & \includegraphics[width=\imguiidsize]{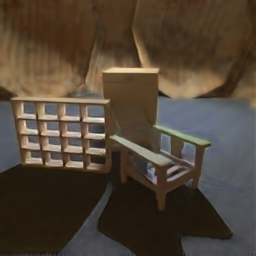} & \includegraphics[width=\imguiidsize]{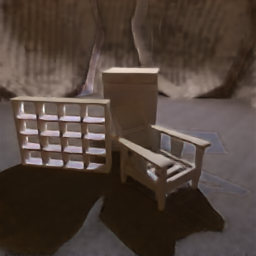} & \includegraphics[width=\imguiidsize]{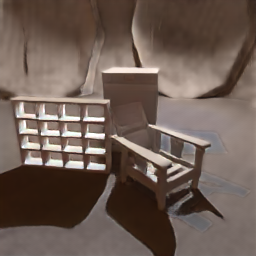} & \includegraphics[width=\imguiidsize]{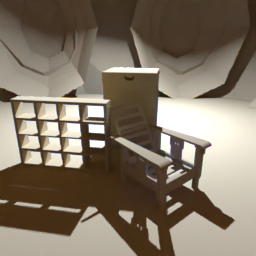} \\
\hspace{-3pt}\rotatebox{90}{\small{\quad \quad Relit}} & \includegraphics[width=\imguiidsize]{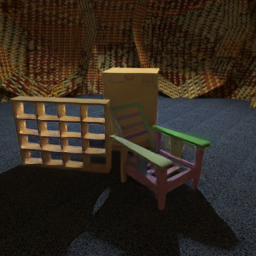} & \includegraphics[width=\imguiidsize]{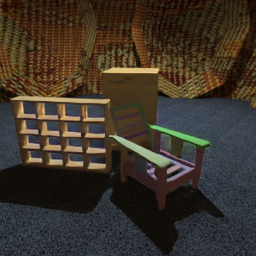} & \includegraphics[width=\imguiidsize]{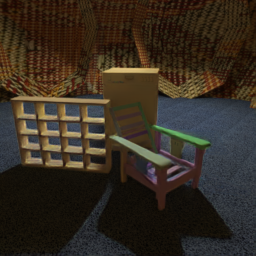} & \includegraphics[width=\imguiidsize]{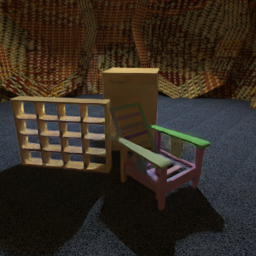} & \includegraphics[width=\imguiidsize]{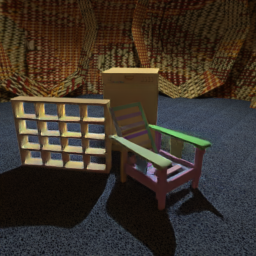} & \includegraphics[width=\imguiidsize]{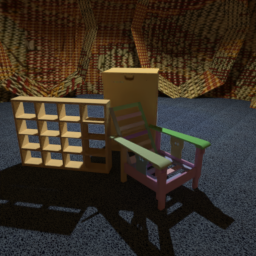} \\
      & w/o $L_{rc}$ & w/o $L_{sc}$ & w/o $L_{scs}+L_{scs}^*$  & w/o $L_{ir}$   & w/ full UIID & GT
\end{tabular}
    \setlength{\tabcolsep}{1pt}
    
    \end{minipage}
\end{figure*}

\paragraph{Comparison with other methods}

\setlength{\tabcolsep}{4pt}
\begin{table}
\begin{center}
\caption{Compare with other methods on ISR dataset}
\label{table:Compare with other methods on ISR dataset}
\begin{tabular}{ccccc}
\hline
                                   & MPS $\uparrow$    & SSIM $\uparrow$   & LPIPS $\downarrow$  & PSNR $\uparrow$  \\ \hline
Pix2pix\cite{isola2017image} & 0.8430                  & 0.8300                   & 0.1439                    & 21.86                    \\
DRN\cite{wang2020deep}     & 0.8925                  & 0.8890                   & 0.1040                    & 24.13                    \\
IAN\cite{zhu2022ian}     & 0.8727                  & 0.8816                   & 0.1361                    & 23.31                    \\
Ours    & \textbf{0.9225}         & \textbf{0.9151}          & \textbf{0.0700}           & \textbf{25.74}  \\ \hline
\end{tabular}
\end{center}
\vspace{-20pt}
\end{table}
\setlength{\tabcolsep}{1.4pt}

We compare our method with three previous methods, Pix2Pix \cite{isola2017image}, DRN \cite{dherse2020scene} and IAN \cite{zhu2022ian}, which were adapted to our single image relighting task. We trained all of them on our ISR dataset. The results are shown in Table \ref{table:Compare with other methods on ISR dataset} and Fig.~\ref{fig: Qualitative comparison with other methods on the ISR dataset.}. Table \ref{table:Compare with other methods on ISR dataset} demonstrates that our method outperforms all of them with a significant margin for all the metrics. In Fig.~\ref{fig: Qualitative comparison with other methods on the ISR dataset.}, we can see that our approach also achieves favorable results in terms of appearance. It is challenging for other methods to accurately predict shadows and obtain high-quality outcomes.

For potential requirements, we provide the parameter quantities and computational costs of the different methods used in our comparison, which are listed in Table \ref{table: The number of parameters and computational costs of each method}. The number of parameters (M) and multiply-accumulate operations (MACs) of our method are similar to those of DRN \cite{wang2020deep}.

\setlength{\tabcolsep}{4pt}
\vspace{-12pt}
\begin{table}[h!]
\begin{center}
\caption{The number of parameters and computational costs of each method}
\vspace{-5pt}
\label{table: The number of parameters and computational costs of each method}
\begin{tabular}{ccccc}
\hline
& Number of parameters(M)  &   \begin{tabular}[c]{@{}l@{}}Multiply-Accumulate\\ Operations(MACs, G)\end{tabular}  \\ \hline
Pix2pix\cite{isola2017image} & 57.5	& 21.4                    \\
DRN\cite{wang2020deep}     & 29.6	& 99.9                    \\
IAN\cite{zhu2022ian}     & 3.8	& 13.1                    \\
Ours    & 23.0 &	102.3 \\ \hline
\end{tabular}
\end{center}
\vspace{-25pt}
\end{table}
\setlength{\tabcolsep}{4pt}

\subsection{Results on the RSR dataset}
\vspace{-2pt}
\setlength{\tabcolsep}{2.5pt}
\begin{table}
\vspace{-5pt}
\begin{center}
\caption{Quantitative results on the RSR dataset}
\label{table:Quantitative results on the RSR dataset}
\begin{tabular}{cccccc}
 \hline
\multicolumn{1}{l}{}     & Pre-trained on ISR & MPS $\uparrow$    & SSIM $\uparrow$   & LPIPS $\downarrow$  & PSNR $\uparrow$  \\ \hline
\multirow{2}{*}{Pix2pix\cite{isola2017image}} & No                 & 0.7938          & 0.7628          & 0.1751          & 19.96          \\
                         & Yes                & 0.8091          & 0.7746          & 0.1563          & 20.26          \\
\multirow{2}{*}{DRN\cite{wang2020deep}}     & No                 & 0.8321          & 0.8113          & 0.1470          & 20.20          \\
                         & Yes                & 0.8691          & 0.8477          & 0.1095          & 22.33          \\
\multirow{2}{*}{IAN\cite{zhu2022ian}}     & No                 & 0.8599          & 0.8492          & 0.1294          & 21.60          \\
                         & Yes                & 0.8676          & 0.8567          & 0.1215          & 22.34          \\
\multirow{2}{*}{Ours}    & No                 & 0.8897          & 0.8655          & 0.0861          & 23.35          \\
                         & Yes                & \textbf{0.9122} & \textbf{0.8879} & \textbf{0.0635} & \textbf{24.63}  \\  \hline
\end{tabular}
\end{center}
\end{table}
\setlength{\tabcolsep}{2.5pt}

\begin{figure*}
\centering
\includegraphics[width=0.95\linewidth]{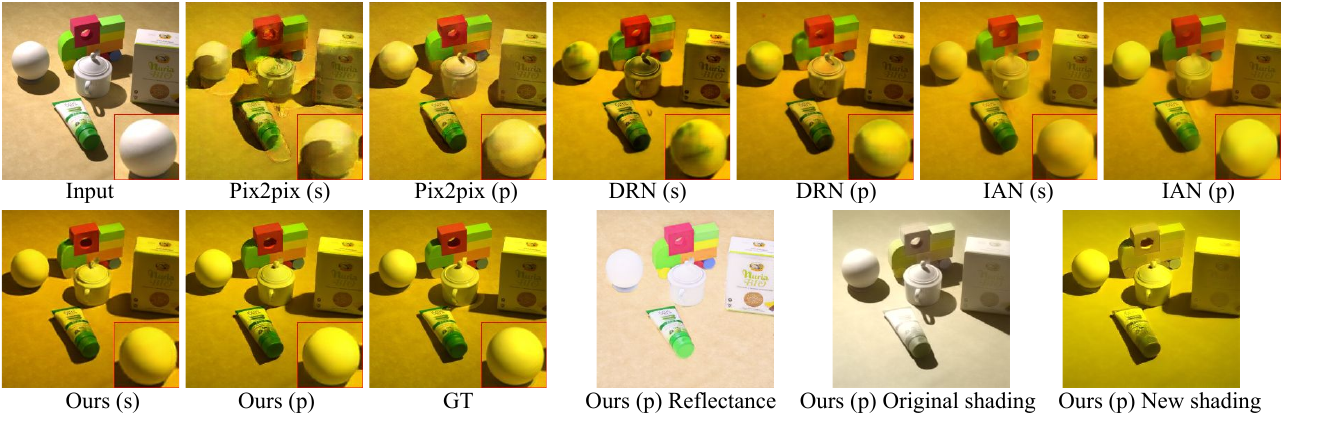}
\vspace{-15pt}
\caption{Qualitative results on the RSR dataset. (s) means: training from scratch on the RSR; (p) means: pretrained on the ISR and fine-tuning on the RSR. The white ball in the image is magnified and placed in the lower right corner for a better comparison.}
\label{fig: Qualitative results on the RSR dataset.}
\vspace{-10pt}
\end{figure*}

In this section, we conduct training and evaluation of our method on the RSR dataset, which is the real scene dataset we propose in section \ref{subsec:RSR dataset}. Additionally, we trained the previous methods using the same settings for comparison. Specifically, in addition to training from scratch, for all methods, we utilize pretrained models from the ISR dataset and fine-tune them on the RSR dataset. The quantitative results are presented in Table \ref{table:Quantitative results on the RSR dataset}, and the qualitative results are depicted in Fig.~\ref{fig: Qualitative results on the RSR dataset.}.

The table clearly demonstrates that our method trained on the RSR dataset outperforms other approaches in all the metrics. Furthermore, the results of all methods indicate that conducting pretraining on the ISR dataset leads to highly significant improvements in the quantitative results. Similar conclusions can be derived from Fig.~\ref{fig: Qualitative results on the RSR dataset.}. In this example, the light condition is transformed from northwest to east with a different light color. Our method achieves the highest accuracy and superior visual appearance. Additionally, Fig.~\ref{fig: Qualitative results on the RSR dataset.} shows the intrinsic components obtained by our method, demonstrating that our approach can successfully achieve reasonable intrinsic decomposition on this real RSR dataset.
\vspace{-8pt}
\subsection{Results on the VIDIT dataset}

\setlength{\tabcolsep}{2.5pt}
\begin{table}
\begin{center}
\caption{Quantitative results on the VIDIT dataset with the setting "any-to-any".}
\label{table: Quantitative results on the VIDIT dataset with the setting "any-to-any"}
\vspace{-6pt}
\begin{tabular}{cccccc}
 \hline
\multicolumn{1}{l}{}     & Pre-trained on ISR & MPS $\uparrow$    & SSIM $\uparrow$   & LPIPS $\downarrow$  & PSNR $\uparrow$  \\ \hline

\multirow{2}{*}{Pix2pix\cite{isola2017image}} & No                 & 0.7236          & 0.6883          & 0.2412          & 18.40          \\
                         & Yes                & 0.7360          & 0.6974          & 0.2255          & 19.07          \\
\multirow{2}{*}{DRN\cite{wang2020deep}}     & No                 & 0.7498          & 0.7213          & 0.2216          & 19.21          \\
                         & Yes                & 0.7646          & 0.7384          & 0.2092          & 19.50          \\
\multirow{2}{*}{IAN\cite{zhu2022ian}}     & No                 & 0.7632          & 0.7410          & 0.2145          & 19.32          \\
                         & Yes                & 0.7776          & 0.7567          & 0.2015          & 19.92          \\
\multirow{2}{*}{Ours}    & No                 & 0.8018          & 0.7704          & 0.1668          & 20.38          \\
                         & Yes                & \textbf{0.8239} & \textbf{0.7967} & \textbf{0.1489} & \textbf{21.50}
 \\ \hline
\end{tabular}
\end{center}
\vspace{-20pt}
\end{table}
\setlength{\tabcolsep}{2.5pt}

\setlength{\tabcolsep}{2.5pt}
\begin{table}
\begin{center}
\caption{Quantitative results on the VIDIT dataset with the setting "one-to-one"}
\label{table: Quantitative results on the VIDIT dataset with the setting "one-to-one"}
\vspace{-6pt}
\begin{tabular}{ccccc}
\hline
\multicolumn{1}{l}{}                & MPS $\uparrow$    & SSIM $\uparrow$   & LPIPS $\downarrow$  & PSNR $\uparrow$  \\ \hline
CET\_SP \cite{helou2020aim}        & 0.6452 & 0.6310 & 0.3405 & 17.07 \\
CET\_CVLab \cite{helou2020aim} & 0.6451 & 0.6362 & 0.3460 & 16.89 \\
DeepRelight \cite{helou2020aim}     & 0.5892 & 0.5928 & 0.4144 & 17.43 \\
DRN \cite{wang2020deep}                                & 0.5780 & 0.5960 & 0.4400 & 17.59 \\
Kubiak \textit{et al.} \cite{kubiak2021silt}          & n/a    & 0.6060 & n/a    & 17.00 \\
IAN \cite{zhu2022ian}                & \textbf{0.6892} & \textbf{0.6861} & 0.3077 & 18.27 \\
Ours                                & 0.6752 & 0.6464 & \textbf{0.2960} & \textbf{18.64} \\
\hline
\end{tabular}
\end{center}
\vspace{-20pt}
\end{table}
\setlength{\tabcolsep}{2.5pt}

\begin{figure*}
\centering
\includegraphics[width=1.0\linewidth]{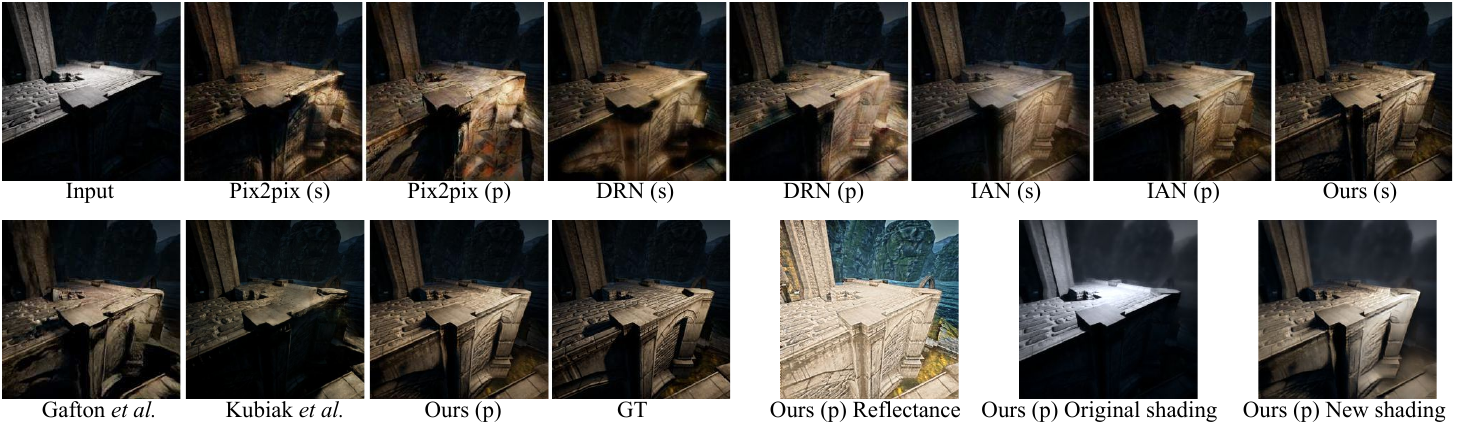}
\vspace{-25pt}
\caption{Qualitative results on the VIDIT dataset (any-to-any). (s) means: training from scratch on VIDIT, (p) means: pretrained on ISR and fine-tuning on VIDIT.}
\vspace{-8pt}
\label{fig: Qualitative results on the VIDIT dataset. any2any}
\end{figure*}

\begin{figure*}[htbp]
   \begin{minipage}[c]{0.84\textwidth}
    \includegraphics[width=1.0\linewidth]{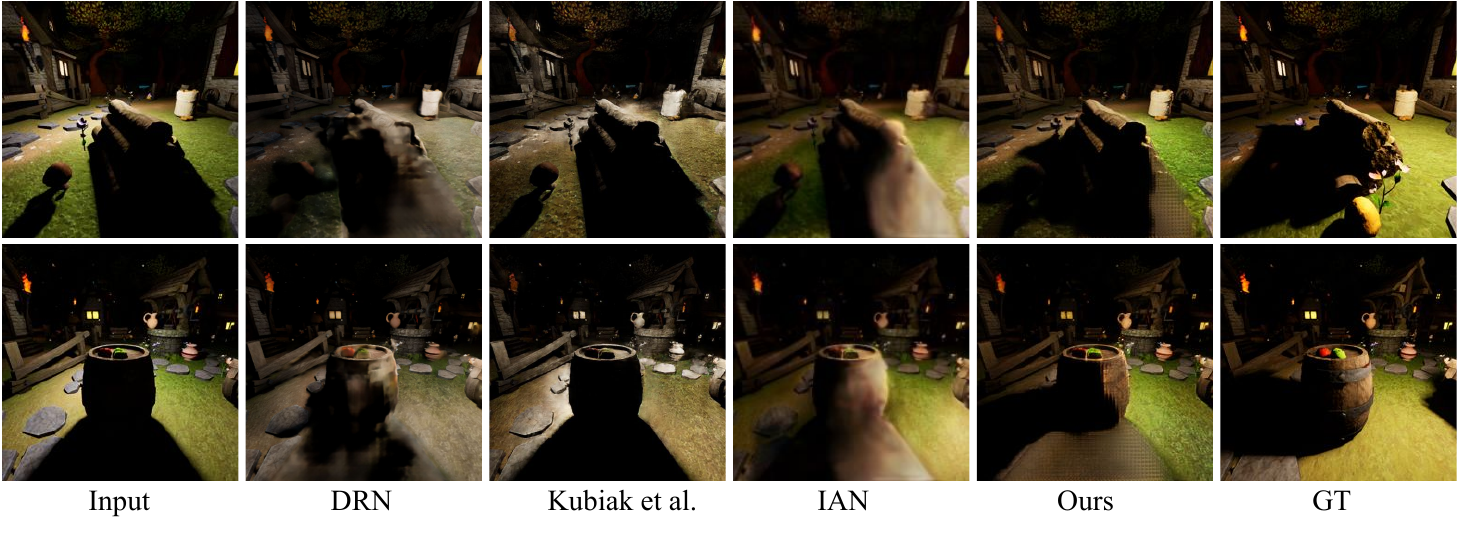}
  \end{minipage}\hfill
  \begin{minipage}[c]{0.15\textwidth}
    \caption{Qualitative comparison with other methods on the VIDIT dataset (one-to-one). All the results of other methods are provided by their official papers or codes. } 
    \label{fig: Qualitative comparison with other methods on the VIDIT dataset (one-to-one).}
  \end{minipage}
   \vspace{-23pt}
\end{figure*}

Since the VIDIT challenge organizers do not disclose the test set and the full validation set, different papers have conducted different experiments and used different train-test set splits. With regard to this diversity, two groups of experiments have been reported. The first group considers "any light", which implies that the options for the input or target lights can vary. This approach has been adopted by Gafton \textit{et al.} \cite{gafton20202d} and the first setting of Kubiak \textit{et al.} \cite{kubiak2021silt}. Specifically, Gafton \textit{et al.} \cite{gafton20202d} explored "any-to-any", whereas Kubiak \textit{et al.} \cite{kubiak2021silt} worked on "any-to-one", which means the target light condition in \cite{kubiak2021silt} was fixed. These works divide the original training set of VIDIT, the only part that contains multiple light conditions, into training and testing sets at a 90:10 ratio. The second group is "one-to-one", which implies a transition from a specific input light condition to a specific output light condition. This case follows the setup provided by track 1 of the challenge \cite{helou2020aim}, which consists of training on the VIDIT training set, validating on the validation set, and conducting testing. This setup has been followed by \cite{zhu2022ian, wang2020deep} and the second setting of Kubiak \textit{et al.} \cite{kubiak2021silt}. We performed experiments emulating the setups of both groups.

Following the setup of the first group, we carry out "any-to-any" relighting, and our split is train:validation:test at an 85:5:10 ratio, which further splits a portion for validation.  The results of this setting are shown in Table \ref{table: Quantitative results on the VIDIT dataset with the setting "any-to-any"} and Fig.~\ref{fig: Qualitative results on the VIDIT dataset. any2any}. In Table \ref{table: Quantitative results on the VIDIT dataset with the setting "any-to-any"}, we apply a similar approach for RSR, not only training from scratch on VIDIT but also pretraining on ISR and fine-tuning on VIDIT. The methods of Gafton \textit{et al.} \cite{gafton20202d} and Kubiak \textit{et al.} \cite{kubiak2021silt}, we do not compare them quantitatively because they did not share their split and their settings are not identical; however, we compare them qualitatively in Fig.~\ref{fig: Qualitative results on the VIDIT dataset. any2any}. 
Table \ref{table: Quantitative results on the VIDIT dataset with the setting "any-to-any"} shows that our method still achieves the best results on the VIDIT dataset, which is also demonstrated in Fig.~\ref{fig: Qualitative results on the VIDIT dataset. any2any}. Fig.~\ref{fig: Qualitative results on the VIDIT dataset. any2any} presents the results from Gafton \textit{et al.} \cite{gafton20202d} and Kubiak \textit{et al.} \cite{kubiak2021silt}. Notably, in this case, they used input images with the same color temperature as the target, which actually corresponds to a simpler setup. The qualitative results further demonstrate that, compared with other methods, our method provides better and more reasonable performance in generating new cast shadows with a good definition of edges. Additionally, in Fig.~\ref{fig: Qualitative results on the VIDIT dataset. any2any}, we also provide intermediate results of our method, which shows that our intrinsic decomposition performs well on the VIDIT dataset.

For the second group with "one-to-one" relighting, we compare with other state-of-the-art methods that have been reported in this experiment. The results are shown in Table \ref{table: Quantitative results on the VIDIT dataset with the setting "one-to-one"} and Fig.~\ref{fig: Qualitative comparison with other methods on the VIDIT dataset (one-to-one).}. 
Table \ref{table: Quantitative results on the VIDIT dataset with the setting "one-to-one"} shows that our method achieves the best LPIPS and PSNR scores. However, regarding the SSIM, it should be noted that our metric is calculated at a resolution of 256$\times$256, whereas the IAN calculates it at a resolution of 1024$\times$1024. We validated the VIDIT image pairs and reported that the SSIM values obtained at a resolution of 256 tend to be 0.03 lower than those at a resolution of 1024. Therefore, our results are still favorable in terms of the SSIM. Furthermore, as the two examples of Fig.~\ref{fig: Qualitative comparison with other methods on the VIDIT dataset (one-to-one).} show, our method is the only one capable of accurately eliminating shadows and generating cast shadows in the correct direction. 

\vspace{-5pt}
\subsection{Results on the Multi-illumination dataset}
\vspace{-2pt}
\setlength{\tabcolsep}{2.5pt}
\begin{table}
\begin{center}
\caption{Quantitative results on the Multi-illumination dataset}
\vspace{-8pt}
\label{table:Quantitative results on the Multi-illumination dataset}
\begin{tabular}{cccccc}
 \hline
\multicolumn{1}{l}{}     & Pre-trained on ISR & MPS $\uparrow$    & SSIM $\uparrow$   & LPIPS $\downarrow$  & PSNR $\uparrow$  \\ \hline

\multirow{2}{*}{Pix2pix\cite{isola2017image}} & No                 & 0.7146          & 0.6560          & 0.2267          & 17.50          \\
                         & Yes                & 0.7037          & 0.6454          & 0.2380          & 18.07          \\
\multirow{2}{*}{DRN\cite{wang2020deep}}     & No                 & 0.7683          & 0.7306          & 0.1940          & 18.96          \\
                         & Yes                & 0.7760          & 0.7404          & 0.1885          & 19.22          \\
\multirow{2}{*}{IAN\cite{zhu2022ian}}     & No                 & 0.7748          & 0.7475          & 0.1979          & 18.94          \\
                         & Yes                & 0.7816          & 0.7547          & 0.1915          & 19.35          \\
\multirow{2}{*}{Ours}    & No                 & 0.8095          & 0.7704          & 0.1515          & 20.64          \\
                         & Yes                & \textbf{0.8117} & \textbf{0.7723} & \textbf{0.1490} & \textbf{20.65}
                         
 \\ \hline
\end{tabular}
\end{center}
\vspace{-20pt}
\end{table}
\setlength{\tabcolsep}{2.5pt}

\begin{figure*}[htbp]
   \begin{minipage}[c]{0.84\textwidth}
    \includegraphics[width=1.0\linewidth]{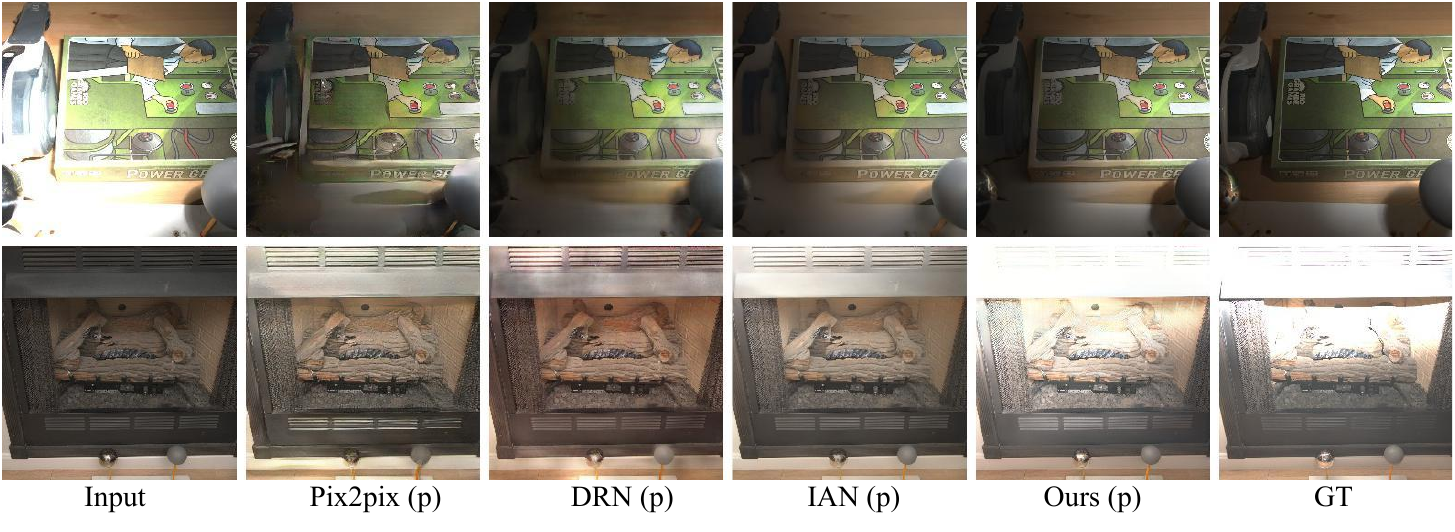}
  \end{minipage}\hfill
  \begin{minipage}[c]{0.15\textwidth}
    \caption{Qualitative comparison with other methods on the Multi-illumination dataset. Owing to space limitations, only the results obtained by each method via pretrained models are provided here. (p) means: pretrained on the ISR and fine-tuned on the Multi-illumination dataset.} 
    \label{fig: Qualitative comparison with other methods on the Multi-illumination dataset}
  \end{minipage}
   \vspace{-21pt}
\end{figure*}

In this section, we extend our experiments to the Multi-illumination dataset, with our focus being on any given input and target light conditions. Although prior methods, such as \cite{kubiak2021silt, zhu2022ian}, have conducted experiments on this dataset, each of their settings is unique and considers only a single target light condition. For a fair comparison, we continue to retrain the aforementioned methods, including \cite{zhu2022ian}. The quantitative and qualitative results are provided in Table \ref{table:Quantitative results on the Multi-illumination dataset} and Fig.~\ref{fig: Qualitative comparison with other methods on the Multi-illumination dataset}, respectively.

As shown in the table, our method consistently yields the best results. However, the benefits of using the pretrained model are not as evident as those in previous experiments, demonstrating pronounced effects only on certain methods and metrics. This diminished impact may be attributed to the substantial differences in light conditions and scene characteristics between the Multi-illumination and ISR datasets. Fig.~\ref{fig: Qualitative comparison with other methods on the Multi-illumination dataset} displays two examples: one demonstrating a transition from light to dark, and the other showing a transition from dark to light. Our method not only produces the most visually pleasing results but also aligns closest to the GT. 
\vspace{-6pt}
\section{Discussion}
The two datasets we introduced provide new dimensions and perspectives for studies in scene relighting. ISR is a large-scale synthetic dataset, whereas RSR provides real-world complexity. Both datasets feature scenes with diverse shading and cast shadows, and provide explicit details regarding light positions and colors, which enables relighting to any target light conditions. These distinctive characteristics set our datasets apart from previous datasets \cite{helou2020vidit, murmann2019dataset} and contribute to mitigating the issue of dataset scarcity in related research. However, even with the addition of our datasets, the complexity of real-world relighting cannot be fully captured.

In the ablation study, we demonstrate the effectiveness of each component in our architecture, including the ResNet backbone, non-local blocks, two-stage model, cross-relighting, and SSIM and LPIPS losses. Additionally, we validate that our method can utilize an unsupervised approach (UIID) to achieve qualitatively pleasing results in intrinsic decomposition, even in the absence of the GT of intrinsic components. Although the results indicate a margin for improvement, we believe that they can still be applied in some downstream tasks, such as image editing \cite{yang2022image}. Experiments on the ISR, RSR, VIDIT \cite{helou2020vidit}, and Multi-Illumination \cite{murmann2019dataset} datasets demonstrate that our method outperforms previous methods. Furthermore, the results indicate that the knowledge learned from the ISR dataset is transferable to other datasets, as evidenced by the experiments with all tested methods. 

\section{Conclusion}
\vspace{-2pt}
In this paper, we introduce two novel datasets, ISR and RSR, for single image scene relighting. This addresses the scarcity of datasets in this research area. Our two-stage network, which integrates intrinsic constraints, achieves superior performance in relighting tasks across various datasets. This network also effectively handles intrinsic decomposition, both in supervised and unsupervised scenarios. The use of physical constraints in decomposition was found to be advantageous for relighting tasks.

Scene relighting from a single image remains a significant challenge with considerable room for improvement. Our future objectives include improving the accuracy and quality of relit images and broadening our relighting ability to encompass a wider variety of scenes, such as images in the wild and of people.

\section*{Acknowledgments}
This work was supported by Grant PID2021-128178OB-I00 funded by MCIN/AEI/10.13039/501100011033 and by ERDF "A way of making Europe". It was also supported by the Departament de Recerca i Universitats from Generalitat de Catalunya, reference 2021SGR01499. We thank Ange Xu for her assistance in generating the synthetic scenes. Yixiong Yang is supported by China Scholarship Council.

\vspace{-6pt}
\bibliographystyle{IEEEtran}
\bibliography{refbib}

\vfill

\end{document}